\documentclass[runningheads]{llncs}

\usepackage{caption} 
\usepackage{color,soulutf8}

\usepackage[hyphens]{url} 
\usepackage{graphicx} 
\usepackage{hyperref}
\usepackage{url}            
\usepackage{amsfonts}       
\usepackage{nicefrac}       
\usepackage{microtype}      
\usepackage{graphicx}
\usepackage{amsmath,amssymb}
\usepackage{bm}
\usepackage{array,booktabs,ragged2e}
\usepackage{cleveref}
\usepackage{subfig}
\usepackage{courier}
\usepackage{longtable}

\usepackage{multicol}
\usepackage[ruled,vlined]{algorithm2e}

\begin{document}

\title{
Personalized Event Prediction \\for Electronic Health Records
}

\titlerunning{Personalized Event Prediction for Electronic Health Records}

\author{Jeong Min Lee\thanks{Work done at University of Pittsburgh, and now affiliated with Meta AI.}\orcidID{0000-0001-8630-0546} \and Milos Hauskrecht\orcidID{0000-0002-7818-0633}}
\institute{Department of Computer Science, University of Pittsburgh, Pittsburgh, PA, USA
\email{\{jlee, milos\}@cs.pitt.edu}}


\maketitle

\begin{abstract} 
    Clinical event sequences consist of hundreds of clinical events that represent records of patient care in time. Developing accurate predictive models of such sequences is of a great importance for supporting a variety of models for interpreting/classifying the current patient condition, or predicting adverse clinical events and outcomes, all aimed to improve patient care. One important challenge of learning predictive models of clinical sequences is their patient-specific variability. Based on underlying clinical conditions, each patient’s sequence may consist of different sets of clinical events (observations, lab results, medications, procedures). Hence, simple population-wide models learned from event sequences for many different patients may not accurately predict patient-specific dynamics of event sequences and their differences. To address the problem, we propose and investigate multiple new event sequence prediction models and methods that let us better adjust the prediction for individual patients and their specific conditions. The methods developed in this work pursue refinement of population-wide models to subpopulations, self-adaptation, and a meta-level model switching that is able to adaptively select the model with the best chance to support the immediate prediction. We analyze and test the performance of these models on clinical event sequences of patients in MIMIC-III database.  
\end{abstract}

\section{Introduction}

Clinical event sequence data based on Electronic Health Records (EHRs) consist of hundreds of clinical events representing records of patient conditions and its management, such as administration of medications, records of lab tests and their results, measurements of various physiological signals, or various procedures. Developing accurate temporal models for such sequences is extremely important for development of various kinds of models defined on clinical data, such as models for predicting adverse events and outcomes \cite{d2021machine,miotto2016deep,tripoliti2017heart,wicki2000predicting,yu2020monitoring}, interpretation of the patient state \cite{lee2019toward,malakouti2019hierarchical,malakouti2019predicting,shamout2019deep,zhang2021interpretable}, understanding the dynamics of the disease and patient condition under different interventions, and/or detection of unusual patient-management actions \cite{hauskrecht2016outlier,hauskrecht_outlier_2013}. All of these may ultimately lead to improved patient care.

One important challenge of learning highly accurate models of clinical sequences is patient-specific variability. Depending on the underlying clinical condition specific to a patient combined with multiple different management options one can choose and apply in patient care, the event patterns may vary widely from patient to patient. Unfortunately, many modern event prediction models and assumptions incorporated into the training of such models may prevent one from accurately representing such a variability. The main challenge, which is also the main topic of this paper, is how to modify or adapt these models to represent better individual patient-specific behaviors and event sequences.  

We study this important challenge in the context of neural autoregressive models. Briefly, neural temporal models based on RNN, LSTM, and attention mechanisms have recently became very popular and widely used for defining and solving various kinds of clinical predictions and representation tasks \cite{choi2016doctor,choi2016multi,choi2016retain,choi2016learning}, including clinical event time-series prediction \cite{lee2022neural,lee2019context,lee_clinical_2020,lee2020multi,lee2021modeling,liu2019nonparametric}. 
However, when these models are built from complex multivariate clinical event sequences, the aforementioned neural models may fail to accurately model patient-specific variability due to their limited ability to represent distributions of dynamic event trajectories. Briefly, the parameters of neural temporal models are learned from many patients data through Stochastic Gradient Descent (SGD) and are shared across all types of patient sequences. 
Hence, the population-wide models tend to average out patient-specific patterns and trajectories in the training sequences. Consequently, they are unable to predict accurately all aspects of patient-specific dynamics of event sequences and their patterns.  

\begin{figure}[t]
\begin{center}

\centerline{\includegraphics[scale=0.48]{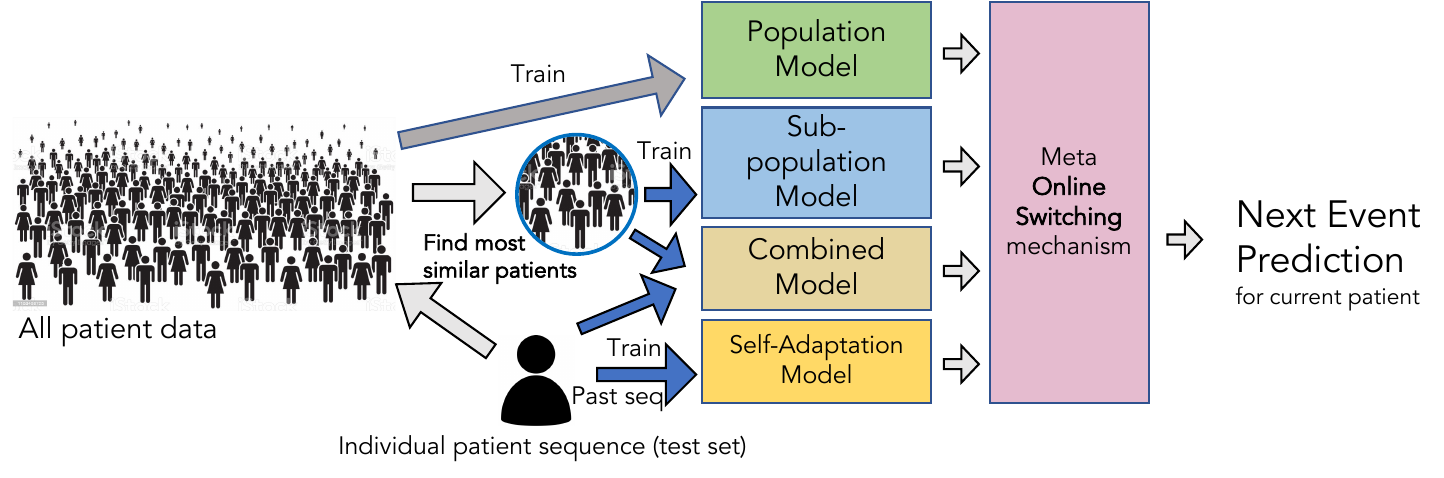}}

\caption{
Overview of approaches we introduced in this work. Along with population model that is trained on all patient data in training set, we explore personalized models that let us better adjust the prediction for individual patient and their specific conditions. The personalized models we study include sub-population model that is trained on the instances of the most similar other patients, self-adaptation model that is trained on the target patient's own past sequence, and combined model that is trained on a dataset combines the two aforementioned data. In this work, we further study meta online switching mechanism that learns to select best performing model's outcome among a pool of models.}

\label{figure:overview}
\end{center}
\end{figure}

To address the above problem, we propose, develop, and study two novel event time-series prediction solutions that attempt to better adapt the population-wide models to the individual patient as shown in \Cref{figure:overview}. First, we propose a model that aims to improve a prediction made for the current patient at any specific time using a repository of event sequences recorded for past patients. The model works by first identifying the patient states among past patients that are most similar to the current state of the current patient and then adapting the predictions of the population-wide model with the help of outcomes recorded for such patients and their states.  We refer to this model as the \emph{subpopulation model}. Second, we develop and study a model that adapts the predictions of the population-wide model only based on the patients' own sequence.  We refer to this model as the {\em self-adaptation model}. However, one concern with either the sub-population or the self-adaptation model and related adaptation is that it may lose some flexibility by being fit too tightly to the specific patient (and patient's recent condition) or to the patient state most similar to the current state.  To address this, we also develop and investigate the meta-switching framework that is able to dynamically identify and switch to the best model to follow for the current patient. Briefly, the meta-framework uses a set of models and learns how to adaptively switch to the model offering the most promising solution. Such a framework may combine the population, subpopulation, and self-adaptation models.
This work extends our previous published work titled ``Neural Clinical Event Sequence Prediction through Personalized Online Adaptive Learning'' {\cite{lee2021neural}}. Based on the foundational personalization methods we studied in the previous work, this work advances adaptive model selecting approaches such as subpopulation-based adaptation, combined adaptation, and meta-level switching mechanisms which greatly increased prediction performance over the methods presented in the previous work.
We note, that all of the above solutions can extend RNN-based multivariate sequence prediction to support personalized clinical event sequence adaptation. We demonstrate the effectiveness of both solutions on clinical event sequences derived from real-world EHRs data from MIMIC-3 Database \cite{johnson2016mimic}. 

\section{Related Work} 
The problem of fitting patient-related outcomes and decisions as close as possible to the target individual has been an essential topic of recent biomedical research and personalized medicine. We briefly list several approaches that build personalized machine learning models for clinical data in the following.

\subsection{Subpopulation Models} 
One classic personalization approach identifies a small set of traits or features that help to define a subpopulation (patient subtype) the patient belongs to, builds a model for the subpopulation, and applies it if a patient from that subpopulation is encountered.  

A straightforward way to define a subpopulation is to use initial clinical observations and demographics. For example, Afrose et al. \cite{afrose2021subpopulation} and Barda et al. \cite{barda2021addressing} create patient subgroups with demographic traits such as race and age. They used the patient subpopulation to solve the data imbalance problem for underrepresented groups in predicting clinical outcomes such as mortality and length of stay. They first learn subpopulation-specific adjustment bias values for calibration purposes. Then, a model's classification outputs are adjusted based on the learned bias values.  

Another approach to defining patient subpopulation is to use clustering methods. For complex clinical data with various types of features, this method has the advantage that it can reveal the latent (hidden) structure and relationship regardless of the complexity of the data representation. In addition, a clustering method can be used for any data representation where the distance metric (or equivalently similarity measure) between data points can be defined. 

Many earlier works on this approach focused on clustering static patient representation such as demographics and symptoms of disease 
\cite{lawton2015parkinson,lewis2005heterogeneity,post2008clinical,van2010identification}. 
More recent work focus on clustering  longitudinal patient representation such as trajectories of biomarkers of kidney function \cite{luong2017k}, opioid usage \cite{mullin2021longitudinal}, or lab test orders \cite{sharafoddini2018finding}. Since this approach considers dynamic changes of clinical features in the data, the discovered patient clusters provide a valuable opportunity for clinical data analysis, such as understanding disease progression \cite{luong2017k,sharafoddini2018finding} or developing more accurate prediction models \cite{mullin2021longitudinal}. For clustering, many earlier works directly use K-means, DBSCAN, or hierarchical clustering algorithms on the top of the features \cite{doshi2014comorbidity,luong2017k,mullin2021longitudinal,sharafoddini2018finding}, and recent works use deep learning based methods to obtain more compact feature representation over the complex clinical data \cite{baytas2017patient,zhang2019data}.

\subsection{Patient-specific Models} 
A more flexible approach to personalized clinical models is to develop patient-specific models that can identify the subpopulation of patients relevant to the target patient by using a patient similarity measure and then build and apply the patient-specific model online whenever the prediction is needed \cite{fojo2017precision,rizopoulos2011dynamic,visweswaran2005instance}. 

One important variability of the clinical time series data is the different sequence lengths. As shown in  {\Cref{figure:number-of-patients}}, while many patients stay in ICU very short amount of time, smaller number of patients stay in ICU longer time. This means little data is available for learning for the patients with longer length of stay. Unnikrishnan and others addressed this issue by building patient-specific models that are trained based on iteratively added most similar other long sequence training instance data {\cite{unnikrishnan2020predicting,unnikrishnan2021love}}. 

Another approach to developing patient-specific models is to use probabilistic sequential latent variable models such as Gaussian Process \cite{schulam2015framework} and Hidden Markov Model \cite{severson20a}. These models have a certain probabilistic form, such as Gaussian distribution for real-valued observations. The parameters for the probability distribution (e.g., mean and variance for Gaussian distribution) are learned during the training process. 
To build a \textit{personalized} probabilistic latent variable model, patient-specific terms are added to the probability distribution parameters. This approach has shown good performance for predicting lab test value (trajectory) of lung disease patients \cite{schulam2015framework} and future complications of Parkinson's disease \cite{severson20a}. 

\subsection{Online Adaptation Methods} However, in many sequential prediction scenarios, the models that are applied to the same patient more than once create an opportunity to adapt and improve the prediction from its past experiences and predictions. This online adaptation lets one improve the patient-specific models and their prediction in time gradually. The standard statistical approach can implement the adaptation process using the Bayesian framework where population-based parameter priors combined with the history of observations and outcomes for the target patient are used to define parameter posteriors \cite{berzuini1992bayesian}. Alternative approaches for online adaptation developed in literature use simpler residual models \cite{liu2016learning_a} that learn the difference (residuals) between the past predictions made by population models and observed outcomes on the current patient. Liu and Hauskrecht \cite{liu2016learning_a} learn these patient-specific residual models for continuous-valued clinical time series and achieve better forecasting performance. 

\subsection{Online Switching Methods} 
The online switching (selection) method is a complementary approach that has been used to increase the prediction performance of online personalization models by allowing multiple (candidate) models to be used together \cite{littlestone1994weighted,shalev2011online}.
At each time in a sequential process, a switching decision is made based on the recent prediction performance of each candidate model. For example, for continuous-valued clinical time series prediction, Liu and Hauskrecht \cite{liu2017personalized} have a pool of population and patient-specific time series models, and at any point in time, the switching method selects the best performing model. 

\subsection{Neural Clinical Event Sequence Prediction} 
EHR-derived clinical event sequence data consists of thousands of sparse and infrequently occurring clinical events. 
In recent years, neural-based models have become the most popular and also the most successful models for representing and predicting EHR-derived clinical sequence data. 
The advantages of such models are their flexibility in modeling latent structures, feature representation, and their learning capability. 
Specifically, word embedding methods 
\cite{mikolov2013distributed}
are effectively used to learn low-dimensional compact representation (embedding) of clinical concepts
\cite{choi2016multi,lee2017diagnosis} and predictive patient state representations \cite{rabhi2022optimized,rabhi2022temporal,tran2015learning}. 
For autoregressive event prediction task, hidden state-space models (e.g., RNN, GRU) and attention mechanism are applied to learn latent dynamics of patient states progression and predict clinical variables such as
diagnosis codes \cite{malakouti2019hierarchical,malakouti2019predicting},
ICU mortality risk \cite{yu2020monitoring}, 
heart failure onset \cite{choi2016retain},
and multivariate future clinical event occurrences \cite{lee2019context,lee_clinical_2020,lee2020multi,lee2021modeling,lee2022learning,liu2019nonparametric}.
For neural-based personalized clinical event prediction, most works focus on using patient-specific feature embedding obtained from patient demographics features \cite{gao2019camp,zhang2018patient2vec}. 
A limitation of the approach is that complex transitions of patient states in time cannot be modeled in a personalized way through static feature embeddings.   
In this work, we develop and investigate methods for adapting modern autoregressive models based on RNN that have been successfully applied to various complex clinical patient states and prediction models.  

\section{Methodology}
\subsection{Neural Autoregressive Event Sequence Prediction}
\label{simple-rnn}
Our goal is to predict occurrences of multiple target events in clinical event sequences. We aim to build an autoregressive model $\phi$ that can predict, at any time $t$, the next step (target) event vector $\bm{y'}_{t+1}$ from a history of past (input) event vectors $\bm{\Theta}_t = \{\bm{y}_{1},\dotsc,\bm{y}_{t}\}$, that is, $\bm{\hat{y}}'_{t+1} = \phi(\bm{\Theta}_t)$. The event vectors are binary $\{0,1\}$ vectors, one dimension per an event type. The input vectors are of dimension $|E|$ where $E$ are different event types in clinical sequences.  The target vector is of dimension $|E'|$, where $E'\subset E$ are events we are interested in predicting. 

One way to build a neural autoregressive prediction model $\phi$ is to use Recurrent Neural Network (RNN) with input embedding matrix $\bm{W}_{emb}$, output linear projection matrix $\bm{W}_{o}$, bias vector $\bm{b}_{o}$, and sigmoid (logit) activation function $\sigma$. At each time step $t$, the RNN-based autoregressive model $\phi$ reads new input $\bm{y}_t$, updates hidden state $\bm{h}_t$, and generates prediction of the target vector $\bm{\hat{y}}'_{t+1}$:

\[ \begin{array}{lll}%
\bm{v}_t = \bm{W}_{emb} \cdot \bm{y}_t &\qquad 
\bm{h}_{t} = \text{RNN}(\bm{h}_{t-1}, \bm{v}_t) &\qquad 
\bm{\hat{y}}'_{t+1} = \sigma(\bm{W}_{o} \cdot \bm{h}_{t} + \bm{b}_{o})\\
\end{array}\]%

$\bm{W}_{emb}, \bm{W}_{o}, \bm{b}_{o}$, and RNN's parameters are learned through SGD with loss function $\mathcal{L}$ defined by the binary cross entropy (BCE):

\begin{equation}
\begin{aligned}
\label{loss}
\mathcal{L} &= \sum_{s \in \mathcal{D}} \sum_{t=1}^{T(s)-1} e(\bm{y'}_{t+1},\bm{\hat{y}'}_{t+1}) \\
\end{aligned}
\end{equation}

\begin{equation}
\begin{aligned}
\label{bce-loss}
e(\bm{y'}_t,\bm{\hat{y}}'_t) &= - [\bm{y'}_{t} \cdot \log \bm{\hat{y}}'_{t} + (\bm{1} - \bm{y'}_{t}) \cdot \log (\bm{1} - \bm{\hat{y}}'_{t})]
\end{aligned}
\end{equation}
where $\mathcal{D}$ is training set and $T(s)$ is length of a sequence $s$. 
This neural autoregressive approach has several benefits when modeling complex high-dimensional clinical sequences: First, low-dimensional embedding with $\bm{W}_{emb}$ helps us to obtain a compact representation of high-dimensional input vector $\bm{y}$. Second, complex dynamics of observed patient state sequences are modeled through RNN which is capable of modeling non-linearities of the sequences. Furthermore, latent variables of neural models typically do not assume a specific probability form. Instead, the complex input-output association is learned through SGD based end-to-end learning framework which allows more flexibility in modeling complex latent dynamics of observed sequence.

However, the neural autoregressive approach cannot address one important characteristic of the clinical sequence: the variability in the dynamics of sequences across different patients. Typically, EHR-derived clinical sequences consist of medical history of several tens of thousands of patients. The dynamics of one patient's sequence could be significantly different from the sequences of other patients. For typical neural autoregressive models, parameters of the trained model are used to process and predict sequences of \textit{all} patients which consist of individual patients who can have different types of clinical complications, medication regimes, or observed sequence dynamics.

\subsection{Subpopulation-based Online Model Adaptation}

To address the patient variability issue, we propose a novel subpopulation-based learning framework that adapts the parameters of the neural autoregressive model to the past patients' sequences that are most similar to the current patient states. 
For simplicity, we denote population model $\phi^{P}$ as a model trained on all training set patient data $\mathcal{D}$, and subpopulation model $\phi^{S}$ as a model that is trained on a subset of training set data $\mathcal{D}^{S}$ that is close to the current patient state. Both models have identical model architecture. 

\noindent \textbf{Non-parametric Memory.}
The proposed learning framework is started by training $\phi^{P}$ with $\mathcal{D}$ and executing inference run for each time step $t' \in T(s')$ of all train set patients $s' \in \mathcal{D}$. Then we define a key-value pair $(k_{t'},v_{t'})$ where the key is the hidden state vector $\bm{h}_{t'}$ and the value is the target event vector $\bm{y}_{t'+1}$. We store $(k_{t'},v_{t'})$ into non-parametric storage (memory) $\mathcal{M}$. 
\begin{equation}
    \mathcal{M} = \{(\bm{h}_{t}, \bm{y}_{t+1})|t \in T(s), s \in \mathcal{D}\}
\end{equation}

\noindent \textbf{Subpopulation Model Initialization.}
Then for each test set patient, we initialize $\phi^{S}$ with the parameters of $\phi^{P}$ to transfer general knowledge about \textit{overall} patient state representation and dynamics to $\phi^{S}$. However, due to patient variability issues, the parameters of $\phi^{P}$ could not be able to fully model the current patient's unique underlying clinical issues and status, and hence its prediction can be limited to correctly predicting the future (next) clinical events. 

\noindent \textbf{Retrieval.}
We approach the aforementioned issue by adapting the parameters of $\phi^{S}$ with additional subpopulation data $\mathcal{D}^{S}$ which will be generated on the fly at each time step $t$ of the current patient sequence.
The subpopulation data $\mathcal{D}^{S}$ is retrieved from $\mathcal{M}$ as a \textit{k}-nearest neighbors $\mathcal{N}$ of the current patient's hidden state $\bm{h}_t$ based on a distance function $d(\cdot,\cdot)$. In this study, we use \textit{L}\textsuperscript{2} distance function which is RBF kernel. The hidden state $\bm{h}_t$ is generated from population model $\phi^{P}$. Since the similarity is calculated on the low-dimensional latent (hidden) state space defined by RNN,  information from both the current input events $\bm{y}_t$ and the dynamics from the series of past events $\bm{y}_1 \dotsc \bm{y}_{t-1}$ is used to compute the similarity between current patient and $\mathcal{M}$. 
\begin{equation}
    \mathcal{D}^{S} = k\text{NN}(\mathcal{M}, \bm{h}_t)
\end{equation}

\noindent \textbf{Subpopulation Model Adaptation.}
We adapt the parameters of $\phi^{S}$ first by computing an subpopulation error $\mathcal{L}^{S} = \sum_{(\bm{h}_i,\bm{y}_{i+1}) \in \mathcal{D}^{S}} e(\bm{y}_{i+1}, \phi^{S}(\bm{h}_i))$. Then, with  $\mathcal{L}^{S}$ we iteratively update parameters of $\phi^{S}$ via SGD. Stopping criterion for the iterative update is: $\mathcal{L}^{S}(\tau-1) - \mathcal{L}^{S}(\tau) < \epsilon$ where $\tau$ denotes the epoch of adaptation update and $\epsilon$ is a positive threshold.

\subsection{Online Self-Adaptation Model}

One limitation of the subpopulation-based model adaptation approach is that we still miss the chance to model the unique dynamics of the current patient's states and its specificity.   
To address this issue, we propose another novel learning framework that adapts the parameters of the neural autoregressive model to the current patient states based on the patient's past event sequence via SGD. We refer to this patient (instance) specific model as $\phi^{I}$. 
As described in Algorithm \ref{algo:online-update}, the online model adaptation procedure at time $t$ for the current patient starts by creating a self-adaptation model $\phi^{I}$ from the population model $\phi^{P}$. Similar to the subpopulation model, $\phi^{I}$ and $\phi^{P}$ have identical model architecture, and values of parameters in $\phi^{I}$ are initialized from $\phi^{P}$ to transfer the knowledge about general representation of patient states and their dynamics. Then, we compute an online patient-specific error $\mathcal{L}^{I}_t = \sum_{i=1}^{t-1} e(\bm{y'}_{i+1},\bm{\hat{y}}'_{i+1})K(t, i)$ that reflects how much the prediction of $\phi^{I}$ deviates from the already observed target sequence for the current patient. 
With $\mathcal{L}^{I}_t$, we iteratively update parameters of $\phi^{I}$ via SGD. The same stopping criterion and training scheme of the subpopulation model is used here for the iterative update of $\phi^{I}$. 

\noindent \textbf{Discounting.}
Please note that our adaptation-based loss $\mathcal{L}^{I}_t$ combines prediction errors for all time steps of the current patient's sequence. However, in order to better fit it to the most recent patient-specific behavior, we weigh the loss more towards recent clinical events. This is done by weighting prediction error for each step $i < t$ with $K(t, i)$ that is based on its time difference from the current time $t$. More specifically, $K(t, i)$ defines an exponential decay function: 

\begin{equation}
\begin{aligned}
\label{eq:exp-kernel}
K(t, i) = \exp{\Big( - \frac{|t - i|}{\gamma} \Big)}
\end{aligned}
\end{equation}
where $\gamma$ denotes the bandwidth (slope) of exponential decay; if $\gamma$ is close to $+\infty$, errors at all time steps have the same weight.

\noindent \textbf{Online Adaptation of Model Components.}
\label{sec:online-component}
The RNN model may have too many parameters, and it may not help to adapt to all of them at the same time. One solution is to relax and permit to adapt only a subset of parameters. On the earlier work on self-adaptation model adaptation \cite{lee2021neural}, three different settings for adapting parameters are experimented and compared: (a) output layer only ($\bm{W}_{o}, \bm{b}_{o}$), (b) transition model (RNN) only, and (c) combination of (a) and (b). From the experiment, (c) adapting only parameters of the output layer showed the best performance for predicting events. Based on this finding, we adapt the parameters of the output layer in this work. 

\begin{algorithm}
    \SetKwInOut{Input}{Input}
    \SetKwInOut{Output}{Output}
\SetAlgoLined
\caption{Online Model Adaptation}
\label{algo:online-update}
\Input{Population model $\phi^{P}$, Current patient's history of \textbf{observed} input sequence $\bm{\Theta}_t=\{\bm{y}_{1},\dotsc,\bm{y}_{t}\}$ and target sequence $(\bm{y'}_{1},\dotsc,\bm{y'}_{t})$}

Initialize self-adaptation model $\phi^{I}$ from $\phi^{P}$; $\tau=0$; $\mathcal{L}^{C}_t(0)=\infty$\;
\Repeat{$\mathcal{L}^{C}_t(\tau-1) - \mathcal{L}^{C}_t(\tau) < \epsilon$}
{
    $\tau = \tau + 1$\;
    $\mathcal{L}^{C}_t(\tau) = \sum_{i=1}^{t-1} e\big(\bm{y'}_{i+1},\bm{\hat{y}}'_{i+1}\big) \cdot K(t, i)$ where $\bm{\hat{y}}'_{i+1}=\phi^{I}(\bm{\Theta}_i)$\; 
    
    Update parameters of $\phi^{I}$ with $\mathcal{L}^{C}_t(\tau)$ via SGD;
}
\Output{Self-adaptation model $\phi^{I}$}
\end{algorithm}

\subsection{Combined Adaptive Model}
The common objective of the two (subpopulation and self-adaptation) adaptation models is to represent better individual patient-specific behaviors and event sequences.  
Indeed, the two models learn different types of information from available patient event sequence data and they are complementary to each other.
By learning from the small pool of most similar past patients' states and its outcome, the subpopulation model can cover dependencies between past and future events which are observed in a small group of patients with specific complications or diseases. On the other hand, the self-adaptation model learns unique dynamics and characteristics of the current patient's own past event sequence.  
Meanwhile, the best way to maximize the gain from the two different approaches is to unify the two methods, and the effective yet straightforward way to unify the two approaches is to combine the two losses $\mathcal{L}^{S}$ and $\mathcal{L}^{I}$ together:
\begin{equation}
    \mathcal{L}^{C} = \mathcal{L}^{I} + \mu \cdot \mathcal{L}^{S}
\end{equation}

In this work, we have the combined adaptation model ${\phi}^{C}$ that is trained on $\mathcal{L}^{C}$ and report its performances along with the previous two approaches.

\subsection{Meta Switching Mechanism}
\label{sec:online-switching}

One limitation of the online adaptation approach is that it tries to modify the dynamics to fit more closely to the specifics of each patient's own sequence or other similar patients' sequences. However, when the patient's state changes suddenly due to recent events (e.g., a sudden clinical complication such as sepsis), the parameters of the adapted models ($\phi^{S,I,C}$) may not be able to adapt quickly enough to these changes. In such a case, switching back to a more general population model could be more desirable.   

Model switching framework \cite{liu2017personalized,shalev2011online} can resolve this issue by dynamically switching among a pool of available models such as subpopulation model $\phi^{S}$, self-adaptation model $\phi^{I}$, combined adapted model $\phi^{C}$, and the population model $\phi^{P}$. Driven by the recent performance of models, it can switch to the best performing model at each time step. 
Algorithm \ref{algo:online-switching} implements the model switching idea. Given a trained population model $\phi^P$, online adapted models $\phi^{S,I,C}$ trained via online adaptation, and the current patient's observed sequence, we can compute discounted losses $\mathcal{L}^{P,S,I,C}$ for these models on the past data. By comparing these losses, we select the model that gives the lowest error (averaged over $|E|$ event types) and use it for predicting the next step.  We refer to prediction based on this meta switching mechanism as \textbf{meta-switching}.

A simple yet powerful extension of the meta switching mechanism is to allow selecting the best model for each event type (event-specific meta switching). One restriction of the aforementioned meta switching mechanism is that one single best model is selected at each time step and the model's prediction for the next time step is used as the output of the meta switching mechanism. We relax this restriction by having \textit{per event type} meta switching mechanism. That is, for each event type we select the best model among a pool of all available models based on each model's performance at the previous time step for each specific event type. This method is referred to as \textbf{meta-switching-event}.

\begin{algorithm}
    \SetKwInOut{Input}{Input}
    \SetKwInOut{Output}{Output}
\SetAlgoLined
\Input{$\phi^{P}$, $\phi^{I}$, $\phi^{S}$, $\phi^{C}$ $\bm{\Theta}_t = \{\bm{y}_{1},\dotsc,\bm{y}_{t}\}$,$(\bm{y'}_{1},\dotsc,\bm{y'}_{t})$}

$\mathcal{L}^{I} = \sum_{i=1}^{t} e({\bm{y'}}_{i+1},{\bm{\hat{y'}}}_{i+1}^{I}) \cdot K(t, i)$ where $\bm{\hat{y'}}^{I}_{i+1}={\phi}^{I}(\bm{\Theta}_i)$\; 
$\mathcal{L}^{P} = \sum_{i=1}^{t} e(\bm{y'}_{i+1},\bm{\hat{y'}}_{i+1}^{P} ) \cdot K(t, i)$ where $\bm{\hat{y'}}_{i+1}^{P}=\phi^{P}(\bm{\Theta}_i)$\;
$\mathcal{L}^{S} = \sum_{i=1}^{t} e(\bm{y'}_{i+1},\bm{\hat{y'}}_{i+1}^{S} ) \cdot K(t, i)$ where $\bm{\hat{y'}}_{i+1}^{S}=\phi^{S}(\bm{\Theta}_i)$\;
$\mathcal{L}^{C} = \sum_{i=1}^{t} e(\bm{y'}_{i+1},\bm{\hat{y'}}_{i+1}^{C} ) \cdot K(t, i)$ where $\bm{\hat{y'}}_{i+1}^{C}=\phi^{C}(\bm{\Theta}_i)$\;
$\bm{\hat{y'}}_{t+1} = \bm{\hat{y'}}_{t+1}^{z} \text{ where } z = \arg \min_{z \in {\{I, P, S, C\}}} \big( \mathcal{L}^{z} \big)$

\Output{Prediction at time step $t+1$: $\bm{\hat{y'}}_{t+1}$}
\caption{Meta Model Switching}
\label{algo:online-switching}
\end{algorithm}

\section{Experimental Evaluation}

\subsection{Experiment Setup}

\subsubsection{Clinical Sequence Generation.}  
\begin{figure}
\begin{center}
\centerline{\includegraphics[scale=0.45]{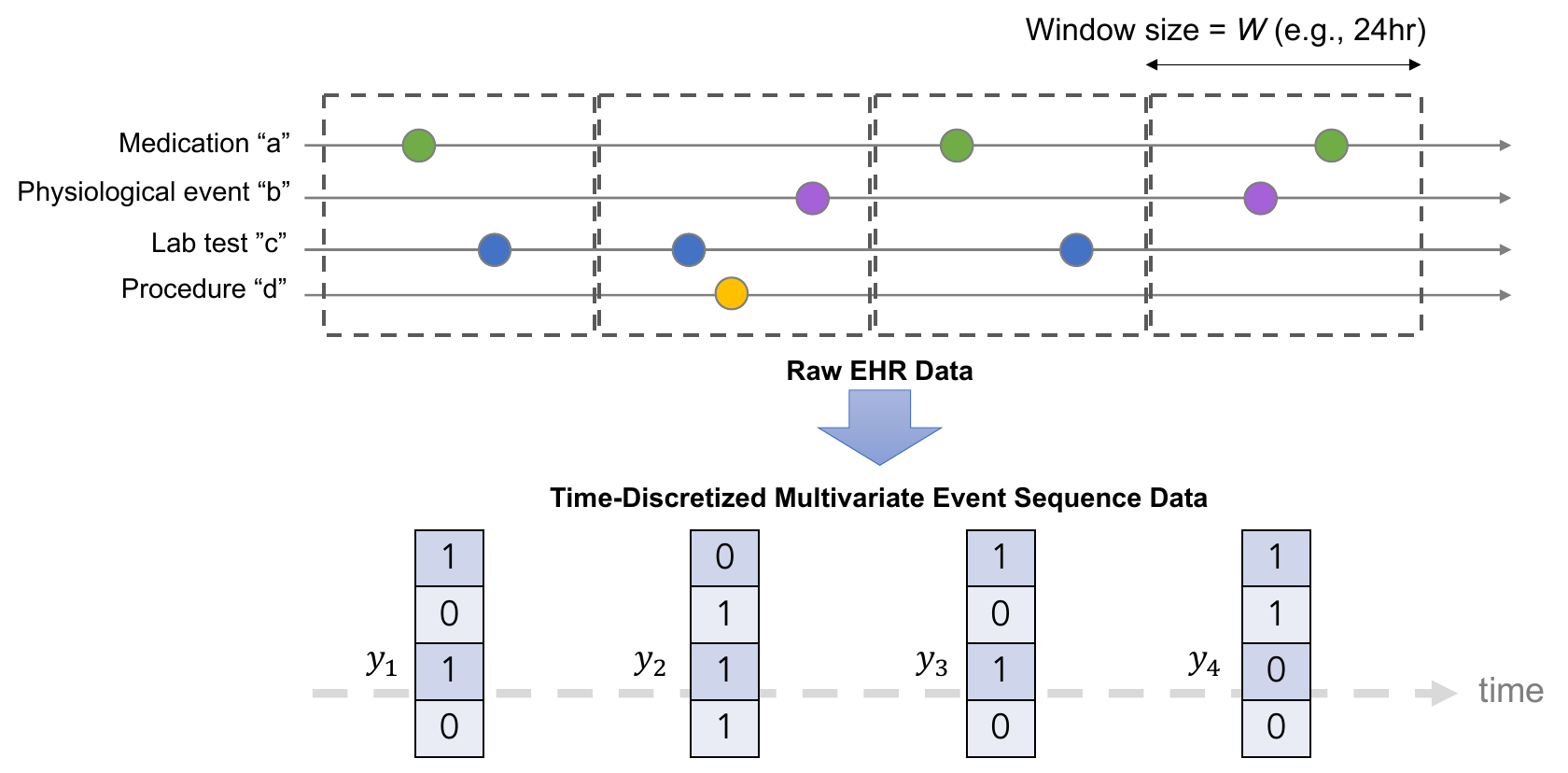}}
\caption{Time discretization of multivariate event sequence data. 
The original EHR-derived clinical time-series data consists of event occurrences in continuous time. We discretized them using a non-overlapping moving window. The events occurred in a single window are represented by a binary vector $y_i \in {{0,1}}^{|E|}$ that cover all event occurrences spanning the period covered by the window.}
\label{win-segmentation}
\end{center}
\end{figure}

From MIMIC-3 \cite{johnson2016mimic}, publicly available EHR database, we extract 5137 patients using the following criteria: (1) patient's age is between 18 and 99, (2) length of admission is between 48 and 480 hours, and (3) clinical records are stored in Meta Vision system, one of the systems used to create MIMIC-3 database. We randomly split the 5137 patients into train and test sets with 80/20 \% split ratio. 
From the extracted records, we generate multivariate event sequences with a sliding-window method. As shown in \Cref{win-segmentation}, we segment the original EHR-derived clinical time-series data in continuous time using a non-overlapping moving window. The events that occurred in a single window of $W$=$24$ hours are represented by a binary vector $\bm{y}_{t} \in {{0,1}}^{|E|}$ that covers all event occurrences spanning the period covered by the window where $t$ denotes a time-step of the window and $E$ is a set of event types. At any point in time $t$, a sequence of vectors created from previous time windows defines an (input) sequence. A vector representing events in the next time window defines the prediction target. 

\noindent \textbf{Feature Extraction.} 
EHR contains thousands of different clinical event types. For efficient modeling we use clinical events that are representative of patient conditions and clinical actions. With this regard, we use four clinical event categories: medication administration events, lab results events, procedure events, and physiological result events. Recent studies in clinical event prediction for EHR data show that using occurrence information (presence/absence) of laboratory tests is more informative than using the measured values of laboratory tests \cite{agniel2018biases,cismondi2013missing,sharafoddini2019new}. Hence, for the lab test and physiological results events, we use occurrence information of each event instead of the values of the observation. 
For medication, lab, and procedure event categories, we filter out those events observed in less than 500 different patients. For physiological events, we select 16 important event types with the help of a critical care physician. 
This process results in 65 medications, 44 procedures, 155 lab tests, and 19 physiological events. The number of the resulting total event ($|E|$) is 283.

\noindent \textbf{Baseline Models.} We compare proposed models to the following baselines: 
\begin{itemize}

\item \textbf{GRU-based POPulation model (GRU-POP)}: For RNN-based time-series modeling described in \Cref{simple-rnn}, we use Gated Recurrent Units (GRU) \cite{chung2014empirical} ($\lambda=$1e-05). With its ability to overcome vanishing gradient issue of RNN, GRU has been widely used in many areas of prediction and modeling of sequence data such as time series {\cite{feng2022multi,tan2020data}}, speech {\cite{graves2014towards,ravanelli2017improving}}, and language {\cite{sutskever2014sequence}} problems and many others. For this reason, we choose GRU as a foundational sequence modeling architecture in this work.
The proposed self-adaptation model (\textbf{SelfAdapt}), Sub-population adaptation model (\textbf{SubpopAdap}), and Combined adaptation model (\textbf{CombinedAdap}) have the same architecture as the population model.
\item \textbf{REverse-Time AttenTioN (RETAIN)}: RETAIN is a representative work on using attention mechanism to summarize clinical event sequences, proposed by Choi et al. \cite{choi2016retain}. It uses two attention mechanisms to comprehend the history of GRU-based hidden states in reverse-time order. For multi-label output, we use a sigmoid function at the output layer.  ($\lambda=$1e-05)
\item \textbf{Logistic regression based on Convolutional Neural Network (CNN)}: 
This model uses CNN to build predictive features summarizing the event history of patients. Following Nguyen et al. \cite{nguyen2016mathtt}, we implement this CNN-based model with a 1-dimensional convolution kernel followed by ReLU activation and max-pooling operation. 
To give more flexibility to the convolution operation, we use multiple kernels with different sizes (2,4,8) and features from these kernels are merged at a fully-connected (FC) layer. ($\lambda=$1e-05)
\end{itemize}

\noindent \textbf{Model Parameters.} 
We use embedding dimension $64$, hidden state dimension $512$, for all neural models. The population model, RETAIN, and CNN use learning rate  $0.005$ and adaptive models use $0.005$. To prevent over-fitting, we use L2 weight decay regularization during the training of GRU-POP, RETAIN, and CNN, and the weight $\lambda$ is determined by the internal cross-validation set (range: 1e-04, 1e-05, 1e-06, 1e-07). For the SGD optimizer, we use Adam.
For the early stopping criteria parameter, we set $\epsilon$=1e-04. 
For kernel bandwidth $\gamma$, we use fixed value 3.0.

\noindent \textbf{Evaluation Metric.} We use the area under the precision-recall curve (AUPRC) as the main evaluation metric. AUPRC is known for presenting a more accurate assessment of the performance of models for a highly imbalanced dataset \cite{saito2015precision}.

\subsection{Results on Personalized Adaptive Models vs. Population Model}

\begin{figure}
\begin{center}
\centerline{\includegraphics[scale=0.55]{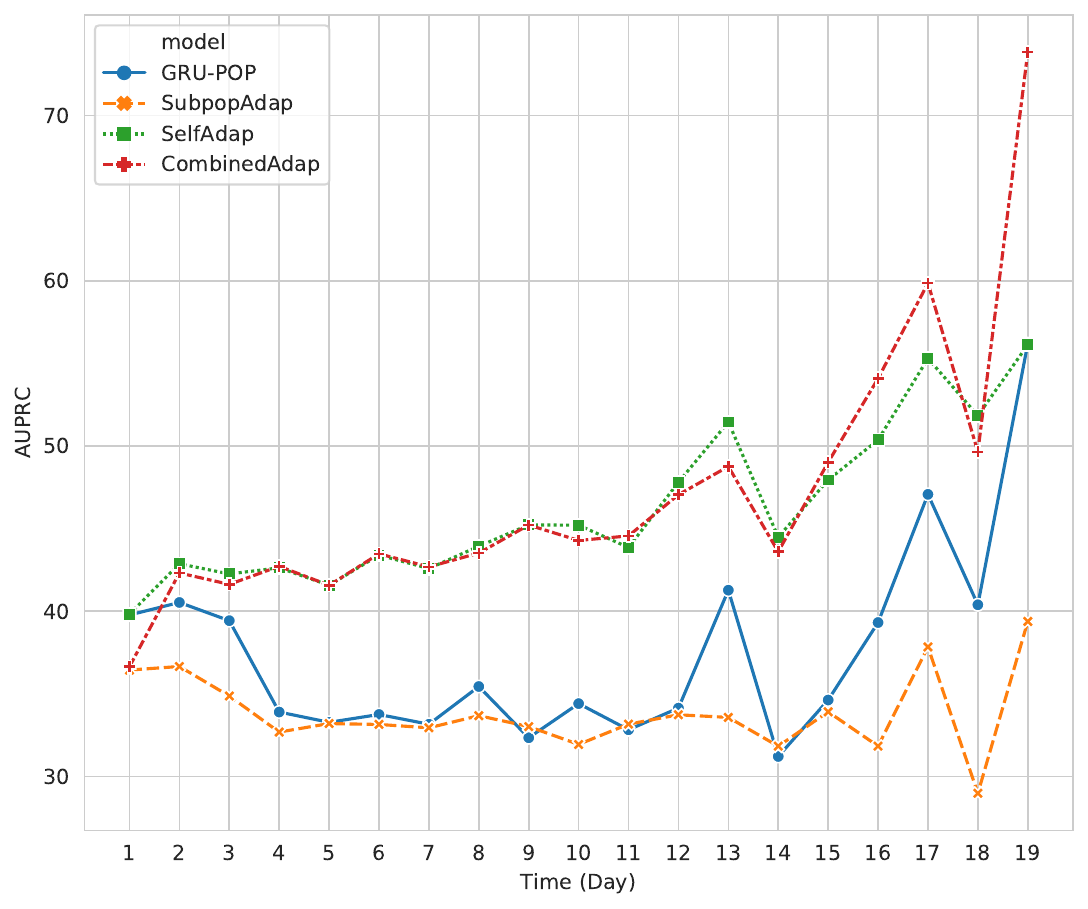}}
\caption{Prediction performance (AUPRC) of the population-based model (GRU-POP) and proposed personalized models based on different mechanisms: subpopulation-based adaptation (SubpopAdap), self-adaptation (SelfAdapt), and combined adaptation (CombinedAdap) which uses both subpopulation and self-adaptation approaches for personalized online adaptation.}
\label{figure:online adaptation}
\end{center}
\end{figure}

We first compare the prediction performance of the population model (GRU-POP) and the proposed methods on different adaptation mechanisms: subpopulation-based adaptation (SubpopAdap), self-adaptation (SelfAdapt), and combined adaptation (CombinedAdap) which uses both subpopulation and self-adaptation approaches for personalized model adaptation. 
As shown in \Cref{figure:online adaptation}, the combined adaptation model and self-adaptation model clearly outperform the population-based model across most of the time steps. Especially on earlier days of admissions (day=1-3), the self-adaptation model performs better than the population model with a decent margin. But subpopulation model underperforms than the population model and it also affects the combined model's performance on the first time step (day) in \Cref{figure:online adaptation}. But as time progresses, the overall performance gap between the combined model and the population model is increasing. On day 19, while the self-adaptation model's performance is almost the same as the population model, the combined model's performance significantly outperform than population model with the help of information from the subpopulation model. That is, we can see that when subpopulation model is solely used, it underperforms than population model overall. This is somehow expected as the parameters of subpopulation model are \textit{indirectly} tuned (adapted) to the current patient through k-nearest neighbor retrieval of other similar patients from the training set data. Therefore, the specificity of the current patient's underlying states is not directly modeled into the parameters of subpopulation model. Nonetheless, the benefit of subpopulation model is revealed through the competency of the combined model. That is, the improved performance of combined model compared to patient specific model can be understood as the additional information provided through the subpopulation model.

\begin{table}
\centering
\small{
\begin{tabular}{lr} 
\toprule
model & AUPRC \\ \midrule
CNN & 37.14 \\ \hline
RETAIN & 34.00 \\ \hline
GRU-POP & 37.54 \\ \hline
SubpopAdap & 33.84 \\ \hline
SelfAdapt & 46.25 \\ \hline
CombinedAdap & 47.08 \\ \hline
Meta-Switch & 49.28 \\ \hline
Meta-Switch-Event & \underline{64.78} \\ \bottomrule
\end{tabular}
}
\caption{Prediction results of all models averaged across all time steps} 

\label{table:result-switching}
\end{table}

\subsection{Results for Meta Switching Mechanism} 

We also experiment with meta online switching approach. It chooses the best predictive model from among a pool of available prediction models. We run the method to choose among the population-based model $\phi^{P}$ and different adaptation models based on subpopulation $\phi^{S}$, self-adaptation approach $\phi^{I}$, and combined approach $\phi^{C}$.

\begin{figure}[t]
\begin{center}

\centerline{\includegraphics[scale=0.55]{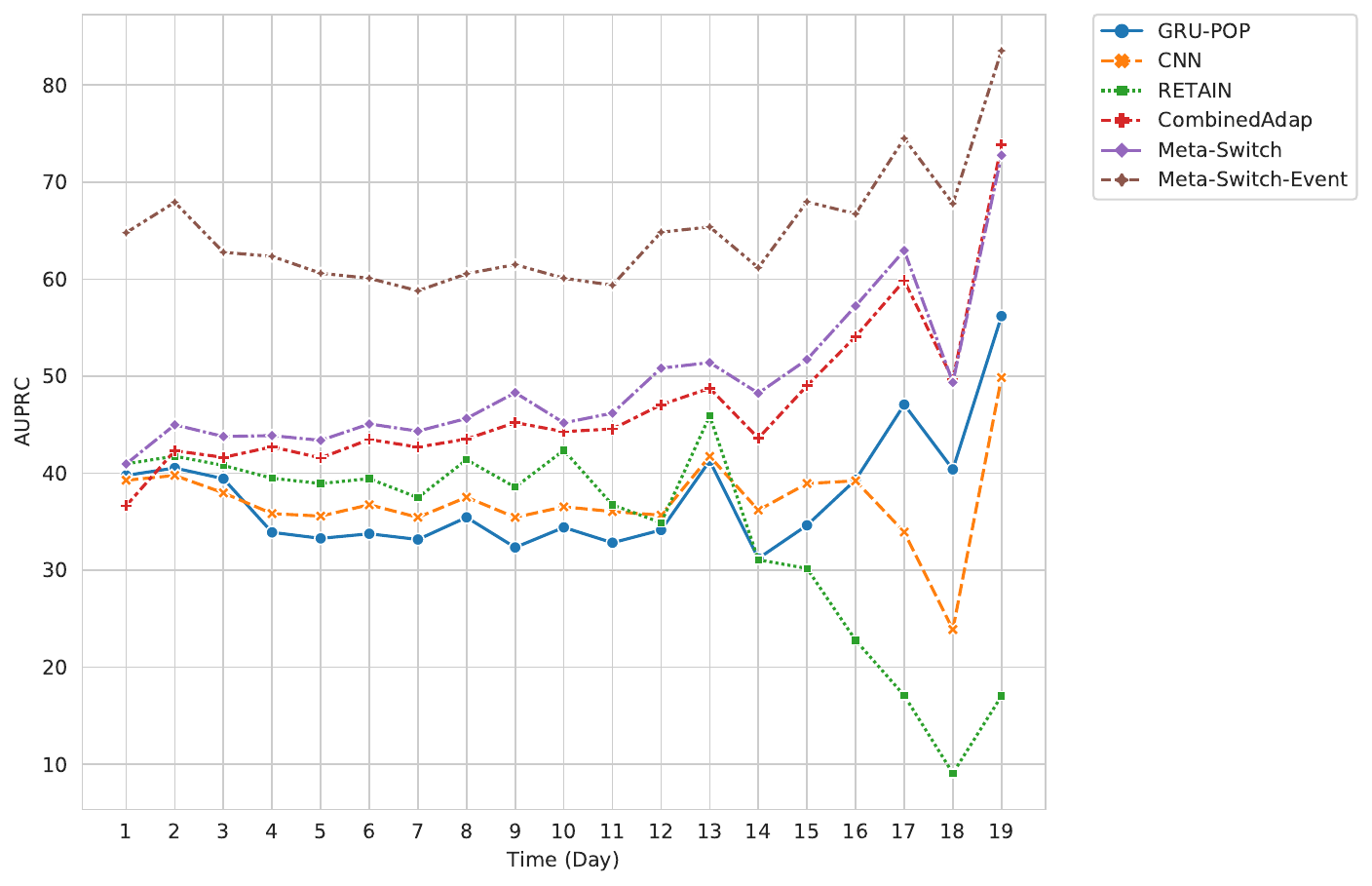}}

\caption{Performance of meta online switching method (Meta-Switch) with population and self-adaptation adaptation models, and its extension to event-specific switching mechanism (Meta-Switch-Event). 
Meta online switching methods 
clearly outperform all baseline models 
(GRU-POP, RETAIN, CNN)
}

\label{figure:online switching}
\end{center}
\end{figure}

As shown in \Cref{figure:online switching}, models that rely on multiple models and online switching clearly outperform baseline models of GRU-POP, CNN, and RETAIN. In particular, the event-specific extension of the meta switching mechanism (Meta-Switch-Event) greatly surpasses the prediction performances of all other models. This shows flexibility in selecting the best model for each event type at each time step substantially benefits the task of predicting complex multivariate clinical event sequence which consists of heterogeneous individual event time series where each has different temporal characteristics and dependencies to precursor events.  

When the prediction performance is averaged across all time steps, we can observe that the event-specific meta-switching mechanism outperforms all models as shown in \Cref{table:result-switching}. Particularly, the event-specific meta switching mechanism's AUPRC is +71\% higher than the population model. The non event-specific version of meta switching increases AUPRC by 31\% from the population model. These results reveal the distinct advantage added by the meta online switching methods.

\begin{figure}[h]
    
    \centering
    \begin{minipage}{0.475\textwidth}
        \centering
        \centerline{\hspace{-2.5mm}\includegraphics[scale=0.55]{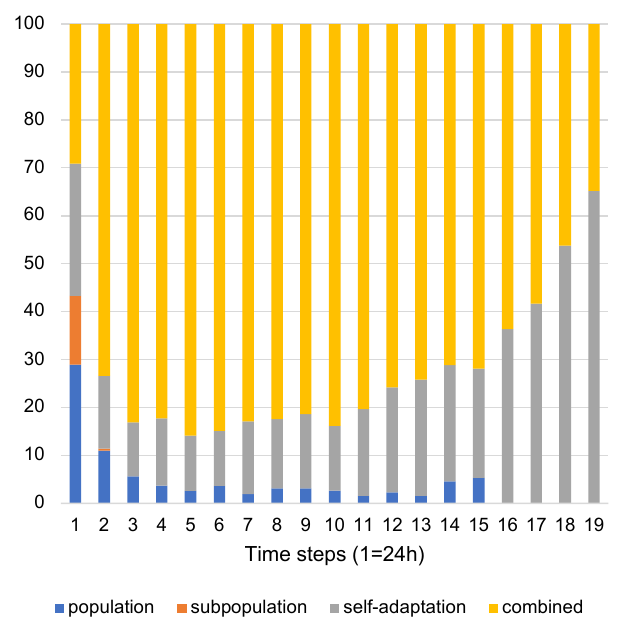}}
        
        \caption{Ratio of different models selected in Meta-Switch mechanism. On earlier time, population and subpopulation models are selected for prediction. On latter time, switching mechanism choose personalized models.}    
        \label{figure:ratio-patient-specific-models}
    \end{minipage}\hfill
    \begin{minipage}{0.475\textwidth}
        \centering
        \centerline{\includegraphics[scale=0.55]{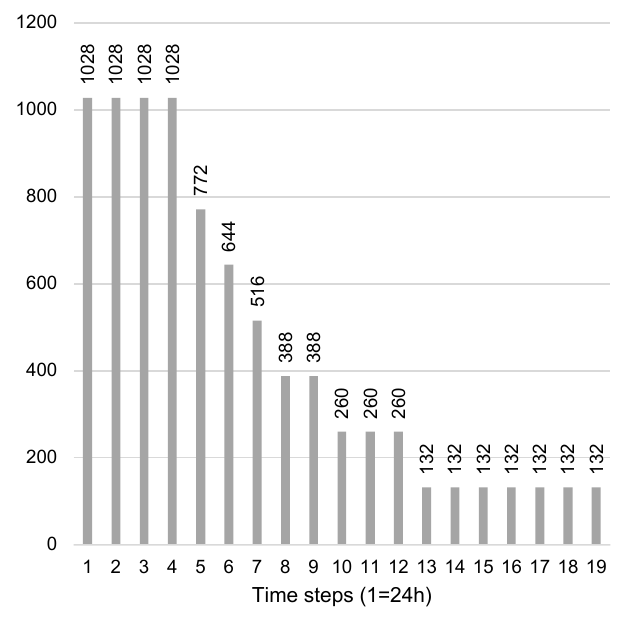}} 
        
        \caption{Number of patients in each time step. The number of patients quickly deteriorates with longer sequence lengths.}        \label{figure:number-of-patients}
    \end{minipage}
    
\end{figure}

\subsubsection{When the model switches?}
To have a better understanding of the behavior of online meta switching-based adaptation, we investigate when the model switches to each model among a pool of available models including the subpopulation model, self-adaptation model, combined model, and population model.
First, we analyze the proportion of how many times each model is used at each time step across all test-set patient sequences from the meta-switching mechanism. As shown in \Cref{figure:ratio-patient-specific-models}, in the first time step, the population model is used 28\% and the subpopulation model is used 14\%. Then, subsequently, the usage ratio of the two models drastically decreased, and the self-adaptation model combined model is mostly used in later time steps. 
Especially, although the direct ratio of the subpopulation model is in general low, its contribution can be found in the fact that the combined model is dominantly used across most time steps (day 2 through day 15). Around the end of the time steps (day 16 through day 19), the ratio for the self-adaptation model is quickly increasing.
This can be explained by the fact that self-adaptation models can have enough observations to adapt the patient-specific variability in that latter time of sequences and it can be a model that provides the best prediction among the pool of all available models. 
To properly interpret the results, \Cref{figure:number-of-patients} shows the number of patients in each time step. This number can also be interpreted as the length of patient sequences and their volume. We can clearly see that the number of patients with longer sequences is very small, as the majority of sequences are very short. For example, patients with sequences longer than 13 days of admission are only about 12\% of all patients in the test set. From this, we can conclude that the population model is often biased towards the dynamics and characteristics of shorter patient sequences. Meanwhile, proposed online adaptation models can effectively learn and adapt better to the dynamics of longer sequences.

\subsubsection{Predicting Repetitive and Non-Repetitive Events.} 
To perform this analysis, we divide event occurrences into two groups based on whether the same type of event has or has not occurred before. We compute AUPRC for each group as shown in \Cref{table:repeat-nonrepeat}.
The results show that for \textbf{non-repetitive events}, the performance of the self-adaptation model is the lowest among all models. This is expected because, with no previous occurrence of a target event, the self-adaptation model could have difficulty making an accurate prediction for the new target event. In this case, we can also see the benefit of the online switching mechanism: the prediction of the population model is more accurate than the self-adaptation model, and the online switching mechanism correctly chooses the population model. More specifically, Meta-Switch mechanism recovers most of the predictability of GRU-POP for non-repetitive event prediction. 
For \textbf{repetitive event prediction}, we can see that both population models and personalized adapted models have similar performances. However, the online switching approaches (Meta-Switch and Meta-Switch-Event) are the best and outperform all other approaches.

\begin{table}[t]

\centering
\small{
\begin{tabular}{lrr} 
\toprule  & Non-repetitive & Repetitive \\ \midrule
CNN & 15.95 & 47.79 \\ \hline
RETAIN & 16.61 & 47.70 \\ \hline
GRU-POP & 16.29 & 48.19 \\ \hline
SubpopAdap & 14.03 & 46.69 \\ \hline
SelfAdapt & 13.00 & 48.17 \\ \hline
CombinedAdap & 14.63 & 48.07 \\ \hline
Meta-Switch & 16.30 & 51.01 \\ \hline
Meta-Switch-Event & \underline{42.93}  & \underline{69.12} \\ \bottomrule
\end{tabular}
}
\caption{Prediction result on non-repetitive and repetitive event groups.
For non-repetitive events, the performance of the self-adaptation model is the lowest. However, the online switching approaches (Meta-Switch, Meta-Switch-Event) recover the predictability by switching to the population model and show the best performance across both groups.
} 

\label{table:repeat-nonrepeat}
\end{table}

\subsubsection{Event-type-specific Performance.} 
\begin{figure}[h]
\begin{center}

\centerline{\includegraphics[scale=0.50]{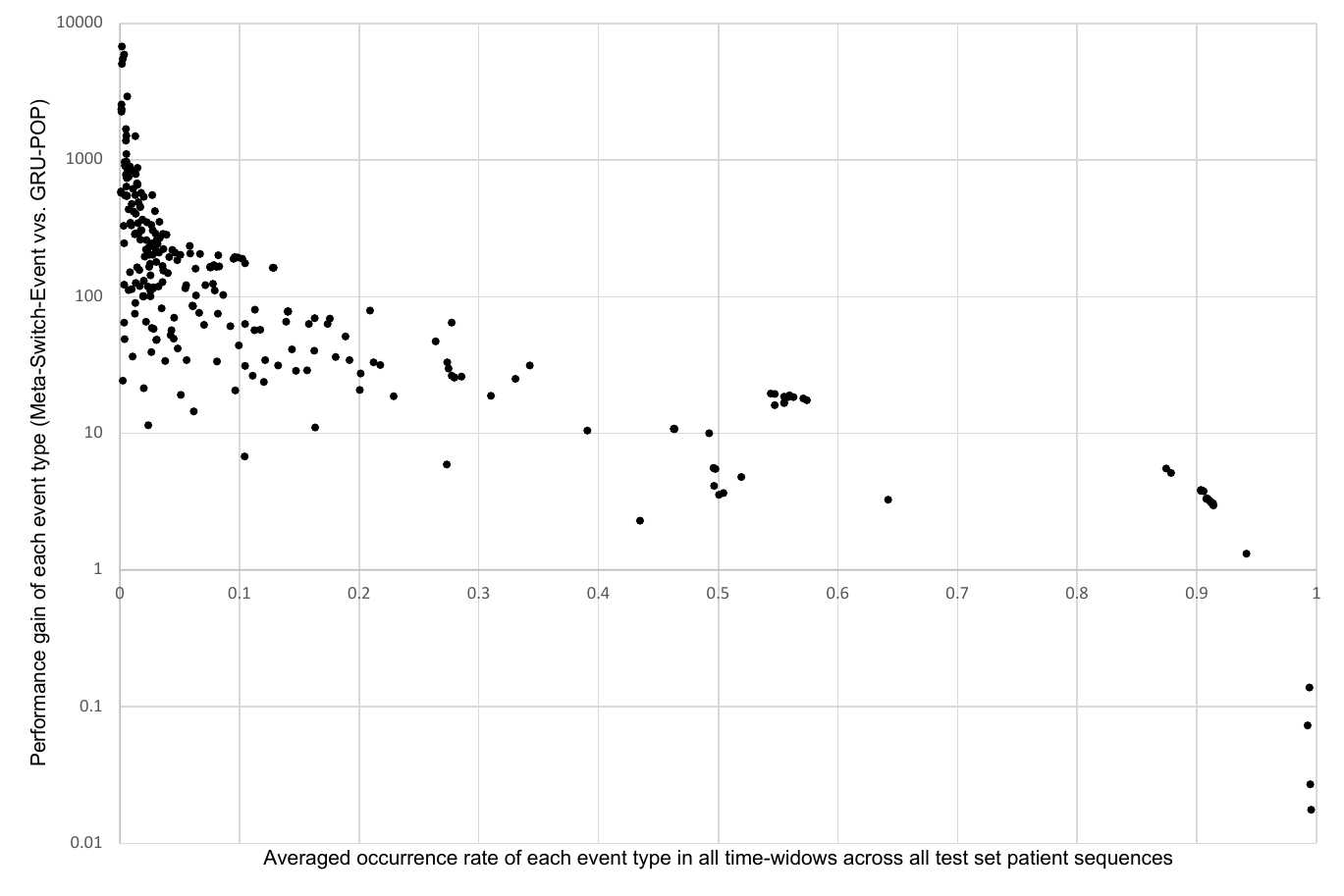}}

\caption{Scatter plot on performance difference between the population model (GRU-POP) and online meta switching-based adaptation model (Meta-Switch-Event) and occurrence rate of each event type.}

\label{figure:scatter-plot}
\end{center}
\end{figure}

We also examine the performance of the online meta switching model (Meta-Switch-Event) compared to the population model (GRU-POP) at the individual event level. Specifically, for each event type, we compute two statistics: first, we compute the percentage difference (\%+) between the two models, and then we compute each event type's occurrence rate in all possible time windows ($W$=$24$), averaged across all test set patient sequences. Then, we plot the two statistics in a scatter plot as shown in \Cref{figure:scatter-plot}. Even the correlation coefficient is weak (-0.24), we can see those event types that have larger performance gaps (e.g., $>$ 100\%+) are indeed less occurring events (e.g., occurrence rate $<$ 0.1). This also reveals that our proposed approaches effectively improve prediction performance, especially for events with smaller data points. It is a valuable characteristic for clinical event time-series prediction where data are usually scarce. The full set of event-specific results can be found in Tables in the Appendix. 

\begin{figure}[h]
\centering
\subfloat[Medication administration\label{fig:2a}] {\hspace{-7mm}\includegraphics[scale=0.61]{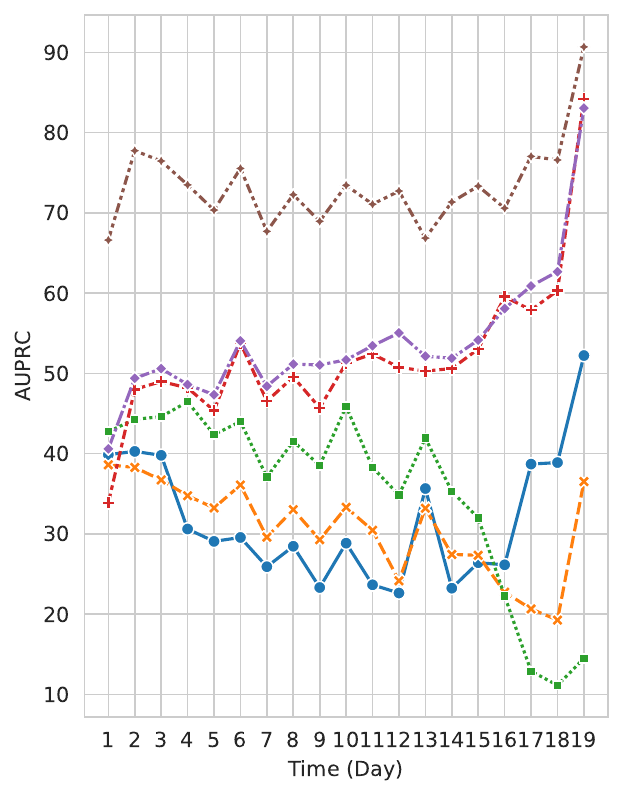}}\hfill 
\subfloat[Lab test results\label{fig:2b}] {\hspace{-3mm}\includegraphics[scale=0.61]{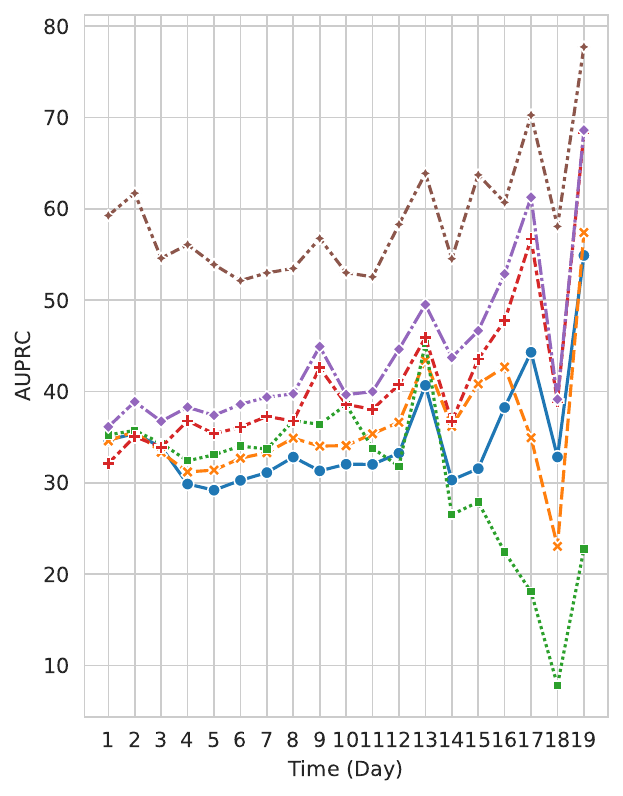}}\hfill
\subfloat[Physiological signal\label{fig:2c}] {\hspace{-7mm}\includegraphics[scale=0.61]{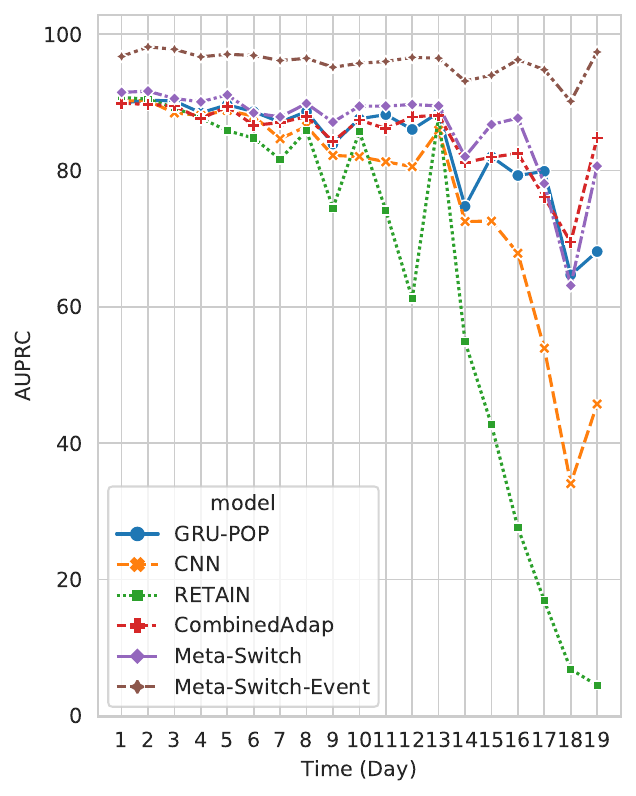}}\hfill
\subfloat[Procedure\label{fig:2d}] {\hspace{-3mm}\includegraphics[scale=0.61]{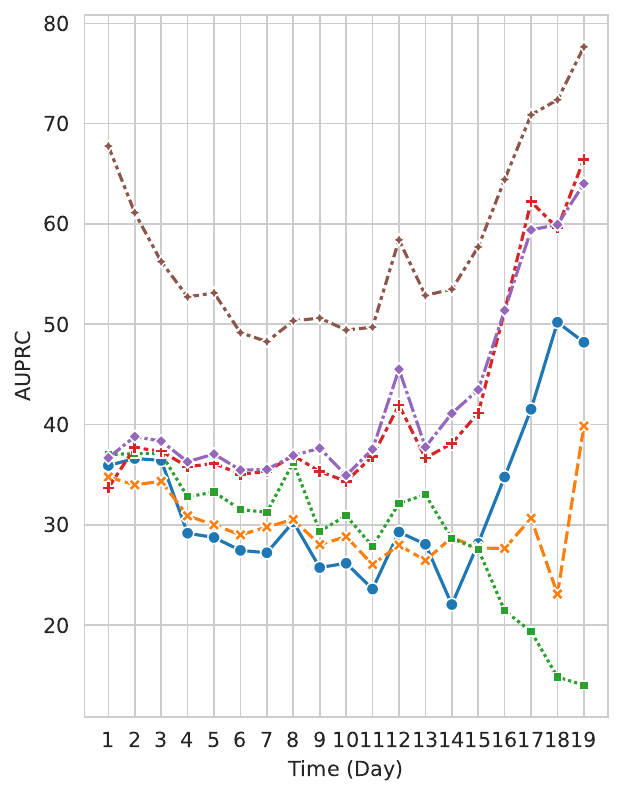}}\hfill

\caption{Prediction results by the event type category}
 \label{figure-event-type}
 \vspace{-12pt}
\end{figure}

\subsubsection{Results based on Event Categories.} 
We analyze the experimental results further into breaking the evaluation results down by inspecting the performances of the models for the four different event categories: medication administration events, lab test events, physiological events, and procedure events. For all $|E|$=$282$ target event types, we averaged prediction performances of them based on the four event categories. The results are shown in Figure \ref{figure-event-type}. Clearly, the proposed methods (Combined Adaptation model, Meta switch mechanism, and Event-specific meta switch mechanism) consistently outperform baseline models across all event categories in AUPRC statistics over all time-steps. Especially, the results of event-specific meta switch mechanism (Meta-Switch-Event) are on a par.

\section{Conclusion}

In this work, we have proposed and investigated multiple new event sequence prediction models and methods that let us better adjust the prediction for individual patients and their specific conditions. The methods developed in this work pursue refinement of population-wide models to subpopulations, self-adaptation, and a meta-level model switching that is able to adaptively select the model with the best chance to support the immediate prediction. These models are of a great importance for defining representations of a patient state and for improving care. We demonstrated the improved performance of our models through experiments on MIMIC-3, a publicly available dataset of electronic health records for ICU patients. 
Nonetheless, to be deployed our work has a few limitations and we want to further explore these limitations in future research. Firstly, these models need to be regularly re-trained to adapt their parameters to dynamically changing patient conditions, and we need to study further about optimal strategy for model update. The too short time interval between two consecutive recurring training sessions may cause instability in the model parameter. The long interval may fail to capture details of patient dynamics and deter such models' efficacy for predictive care. Future studies could investigate dynamic model parameter update, store, and retrieval strategies. 
Secondly, since we create and train individual models for each patient, we need to have a sufficient scalable computing infrastructure to be able to serve thousands of patients in real-time concurrently. 
Overall, while our work has demonstrated promising results, further research is needed to fully evaluate and validate the potential of our proposed models for clinical practice.

\section*{Acknowledgement} 

This work was supported by NIH grants R01-EB032752 and R01-GM088224. The content of this paper is solely the responsibility of the authors and does not necessarily represent the official views of the NIH.

\bibliographystyle{plain} 
\bibliography{BibFile}

\begin{thebibliography}{10}

\bibitem{afrose2021subpopulation}
Sharmin Afrose, Wenjia Song, Charles~B Nemeroff, Chang Lu, and Danfeng~Daphne
  Yao.
\newblock Subpopulation-specific machine learning prognosis for
  underrepresented patients with double prioritized bias correction.
\newblock {\em medRxiv}, 2021.

\bibitem{agniel2018biases}
Denis Agniel, Isaac~S Kohane, and Griffin~M Weber.
\newblock Biases in electronic health record data due to processes within the
  healthcare system: retrospective observational study.
\newblock {\em Bmj}, 361, 2018.

\bibitem{barda2021addressing}
Noam Barda, Gal Yona, Guy~N Rothblum, Philip Greenland, Morton Leibowitz, Ran
  Balicer, Eitan Bachmat, and Noa Dagan.
\newblock Addressing bias in prediction models by improving subpopulation
  calibration.
\newblock {\em Journal of the American Medical Informatics Association},
  28(3):549--558, 2021.

\bibitem{baytas2017patient}
Inci~M Baytas, Cao Xiao, Xi~Zhang, Fei Wang, Anil~K Jain, and Jiayu Zhou.
\newblock Patient subtyping via time-aware lstm networks.
\newblock In {\em Proceedings of the 23rd ACM SIGKDD international conference
  on knowledge discovery and data mining}, pages 65--74, 2017.

\bibitem{berzuini1992bayesian}
Carlo Berzuini, Riccardo Bellazzi, Silvana Quaglini, and D.J. Spiegelhalter.
\newblock Bayesian networks for patient monitoring.
\newblock {\em Artificial Intelligence in Medicine}, 4:243–260, 05 1992.

\bibitem{choi2016doctor}
Edward Choi, Mohammad~Taha Bahadori, Andy Schuetz, Walter~F Stewart, and Jimeng
  Sun.
\newblock Doctor ai: Predicting clinical events via recurrent neural networks.
\newblock In {\em Machine Learning for Healthcare Conference}, pages 301--318,
  2016.

\bibitem{choi2016multi}
Edward Choi, Mohammad~Taha Bahadori, Elizabeth Searles, Catherine Coffey,
  Michael Thompson, James Bost, Javier Tejedor-Sojo, and Jimeng Sun.
\newblock Multi-layer representation learning for medical concepts.
\newblock In {\em The 22nd ACM SIGKDD International Conference on Knowledge
  Discovery and Data Mining}, pages 1495--1504. ACM, 2016.

\bibitem{choi2016retain}
Edward Choi, Mohammad~Taha Bahadori, Jimeng Sun, Joshua Kulas, Andy Schuetz,
  and Walter Stewart.
\newblock Retain: An interpretable predictive model for healthcare using
  reverse time attention mechanism.
\newblock In {\em Advances in Neural Information Processing Systems}, pages
  3504--3512, 2016.

\bibitem{choi2016learning}
Youngduck Choi, Chill Yi-I Chiu, and David Sontag.
\newblock Learning low-dimensional representations of medical concepts.
\newblock {\em AMIA Summits on Translational Science Proceedings}, 2016:41,
  2016.

\bibitem{chung2014empirical}
Junyoung Chung, Caglar Gulcehre, KyungHyun Cho, and Yoshua Bengio.
\newblock Empirical evaluation of gated recurrent neural networks on sequence
  modeling.
\newblock {\em arXiv preprint arXiv:1412.3555}, 2014.

\bibitem{cismondi2013missing}
Federico Cismondi, Andr{\'e}~S Fialho, Susana~M Vieira, Shane~R Reti,
  Jo{\~a}o~MC Sousa, and Stan~N Finkelstein.
\newblock Missing data in medical databases: Impute, delete or classify?
\newblock {\em Artificial intelligence in medicine}, 58(1):63--72, 2013.

\bibitem{d2021machine}
Fabrizio D'Ascenzo, Ovidio De~Filippo, Guglielmo Gallone, Gianluca Mittone,
  Marco~Agostino Deriu, Mario Iannaccone, Albert Ariza-Sol{\'e}, Christoph
  Liebetrau, Sergio Manzano-Fern{\'a}ndez, Giorgio Quadri, et~al.
\newblock Machine learning-based prediction of adverse events following an
  acute coronary syndrome (praise): a modelling study of pooled datasets.
\newblock {\em The Lancet}, 397(10270):199--207, 2021.

\bibitem{doshi2014comorbidity}
Finale Doshi-Velez, Yaorong Ge, and Isaac Kohane.
\newblock Comorbidity clusters in autism spectrum disorders: an electronic
  health record time-series analysis.
\newblock {\em Pediatrics}, 133(1):e54--e63, 2014.

\bibitem{feng2022multi}
Zhong-kai Feng, Qing-qing Huang, Wen-jing Niu, Tao Yang, Jia-yang Wang, and
  Shi-ping Wen.
\newblock Multi-step-ahead solar output time series prediction with gate
  recurrent unit neural network using data decomposition and cooperation search
  algorithm.
\newblock {\em Energy}, 261:125217, 2022.

\bibitem{fojo2017precision}
Anthony~T Fojo, Katherine~L Musliner, Peter~P Zandi, and Scott~L Zeger.
\newblock A precision medicine approach for psychiatric disease based on
  repeated symptom scores.
\newblock {\em Journal of psychiatric research}, 95:147--155, 2017.

\bibitem{gao2019camp}
Jingyue Gao, Xiting Wang, Yasha Wang, Zhao Yang, Junyi Gao, Jiangtao Wang, Wen
  Tang, and Xing Xie.
\newblock Camp: Co-attention memory networks for diagnosis prediction in
  healthcare.
\newblock ICDM, 2019.

\bibitem{graves2014towards}
Alex Graves and Navdeep Jaitly.
\newblock Towards end-to-end speech recognition with recurrent neural networks.
\newblock In {\em International Conference on Machine Learning}, pages
  1764--1772, 2014.

\bibitem{hauskrecht2016outlier}
Milos Hauskrecht, Iyad Batal, Charmgil Hong, Quang Nguyen, Gregory~F Cooper,
  Shyam Visweswaran, and Gilles Clermont.
\newblock Outlier-based detection of unusual patient-management actions: an icu
  study.
\newblock {\em Journal of biomedical informatics}, 64:211--221, 2016.

\bibitem{hauskrecht_outlier_2013}
Milos Hauskrecht, Iyad Batal, Michal Valko, Shyam Visweswaran, Gregory~F
  Cooper, and Gilles Clermont.
\newblock Outlier detection for patient monitoring and alerting.
\newblock {\em Journal of biomedical informatics}, 46(1):47--55, 2013.

\bibitem{johnson2016mimic}
Alistair~EW Johnson, Tom~J Pollard, Lu~Shen, H~Lehman Li-wei, Mengling Feng,
  Mohammad Ghassemi, Benjamin Moody, Peter Szolovits, Leo~Anthony Celi, and
  Roger~G Mark.
\newblock {MIMIC-III}, a freely accessible critical care database.
\newblock {\em Scientific data}, 3:160035, 2016.

\bibitem{lawton2015parkinson}
Michael Lawton, Fahd Baig, Michal Rolinski, Claudio Ruffman, Kannan Nithi,
  Margaret~T May, Yoav Ben-Shlomo, and Michele Hu.
\newblock Parkinson’s disease subtypes in the oxford parkinson disease centre
  (opdc) discovery cohort.
\newblock {\em Journal of Parkinson's disease}, 5(2):269--279, 2015.

\bibitem{lee2019toward}
Eunho Lee, Jun-Sik Choi, Minjeong Kim, Heung-Il Suk, Alzheimer’s
  Disease~Neuroimaging Initiative, et~al.
\newblock Toward an interpretable alzheimer’s disease diagnostic model with
  regional abnormality representation via deep learning.
\newblock {\em Neuroimage}, 202:116113, 2019.

\bibitem{lee2022neural}
Jeong~Min Lee.
\newblock {\em Neural Event Prediction for Clinical Event Time-Series}.
\newblock PhD thesis, University of Pittsburgh, 2022.

\bibitem{lee2019context}
Jeong~Min Lee and Milos Hauskrecht.
\newblock Recent-context-aware lstm-based clinical time-series prediction.
\newblock In {\em In Proceedings of AI in Medicine Europe (AIME)}, 2019.

\bibitem{lee_clinical_2020}
Jeong~Min Lee and Milos Hauskrecht.
\newblock Clinical {Event} {Time}-series {Modeling} with {Periodic} {Events}.
\newblock In {\em The {Thirty}-{Third} {International} {Flairs} {Conference}}.
  AAAI, 2020.

\bibitem{lee2020multi}
Jeong~Min Lee and Milos Hauskrecht.
\newblock Multi-scale temporal memory for clinical event time-series
  prediction.
\newblock In {\em 2020 International Conference on Artificial Intelligence in
  Medicine (AIME 2020)}, 2020.

\bibitem{lee2021modeling}
Jeong~Min Lee and Milos Hauskrecht.
\newblock Modeling multivariate clinical event time-series with recurrent
  temporal mechanisms.
\newblock {\em Artificial Intelligence in Medicine}, 112:102021, 2021.

\bibitem{lee2021neural}
Jeong~Min Lee and Milos Hauskrecht.
\newblock Neural clinical event sequence prediction through personalized online
  adaptive learning.
\newblock In {\em 19th International Conference on Artificial Intelligence in
  Medicine (AIME 2021)}, pages https--arxiv.
  https://link.springer.com/chapter/10.1007\%2F978-3-030-77211-6\_20, 2021.

\bibitem{lee2022learning}
Jeong~Min Lee and Milos Hauskrecht.
\newblock Learning to adapt clinical sequences with residual mixture of
  experts.
\newblock In {\em 2022 International Conference on Artificial Intelligence in
  Medicine (AIME 2022)}, pages https--arxiv, 2022.

\bibitem{lee2017diagnosis}
Jeong~Min Lee and Aldrian~Obaja Muis.
\newblock Diagnosis code prediction from electronic health records as
  multilabel text classification: a survey.
\newblock {\em URL: http://people. cs. pitt. edu/\~{}
  jlee/papers/cp1\_survey\_jlee\_amuis. pdf [accessed 2021-05-09]}, 2017.

\bibitem{lewis2005heterogeneity}
SJG Lewis, Thomas Foltynie, Andrew~D Blackwell, Trevor~W Robbins, Adrian~M
  Owen, and Roger~A Barker.
\newblock Heterogeneity of parkinson’s disease in the early clinical stages
  using a data driven approach.
\newblock {\em Journal of Neurology, Neurosurgery \& Psychiatry},
  76(3):343--348, 2005.

\bibitem{littlestone1994weighted}
Nick Littlestone and Manfred~K. Warmuth.
\newblock The weighted majority algorithm.
\newblock {\em Inf. Comput.}, 108(2):212–261, February 1994.

\bibitem{liu2019nonparametric}
Siqi Liu and Milos Hauskrecht.
\newblock Nonparametric regressive point processes based on conditional
  gaussian processes.
\newblock In {\em Advances in Neural Information Processing Systems}, pages
  1062--1072, 2019.

\bibitem{liu2016learning_a}
Zitao Liu and Milos Hauskrecht.
\newblock Learning adaptive forecasting models from irregularly sampled
  multivariate clinical data.
\newblock In {\em The 30th AAAI Conference on Artificial Intelligence}, pages
  1273--1279, 2016.

\bibitem{liu2017personalized}
Zitao Liu and Milos Hauskrecht.
\newblock A personalized predictive framework for multivariate clinical time
  series via adaptive model selection.
\newblock In {\em Proceedings of the 2017 ACM on Conference on Information and
  Knowledge Management}, pages 1169--1177, 2017.

\bibitem{luong2017k}
Duc Thanh~Anh Luong and Varun Chandola.
\newblock A k-means approach to clustering disease progressions.
\newblock In {\em 2017 IEEE International conference on healthcare informatics
  (ICHI)}, pages 268--274. IEEE, 2017.

\bibitem{malakouti2019hierarchical}
S.~{Malakouti} and M.~{Hauskrecht}.
\newblock Hierarchical adaptive multi-task learning framework for patient
  diagnoses and diagnostic category classification.
\newblock In {\em 2019 IEEE International Conference on Bioinformatics and
  Biomedicine (BIBM)}, pages 701--706, 2019.

\bibitem{malakouti2019predicting}
Seyedsalim Malakouti and Milos Hauskrecht.
\newblock Predicting patient’s diagnoses and diagnostic categories from
  clinical-events in ehr data.
\newblock In {\em Conference on Artificial Intelligence in Medicine in Europe},
  pages 125--130. Springer, 2019.

\bibitem{mikolov2013distributed}
Tomas Mikolov, Ilya Sutskever, Kai Chen, Greg~S Corrado, and Jeff Dean.
\newblock Distributed representations of words and phrases and their
  compositionality.
\newblock In {\em Advances in neural information processing systems}, pages
  3111--3119, 2013.

\bibitem{miotto2016deep}
Riccardo Miotto, Li~Li, Brian~A Kidd, and Joel~T Dudley.
\newblock Deep patient: an unsupervised representation to predict the future of
  patients from the electronic health records.
\newblock {\em Scientific reports}, 6:26094, 2016.

\bibitem{mullin2021longitudinal}
Sarah Mullin, Jaroslaw Zola, Robert Lee, Jinwei Hu, Brianne MacKenzie, Arlen
  Brickman, Gabriel Anaya, Shyamashree Sinha, Angie Li, and Peter~L Elkin.
\newblock Longitudinal k-means approaches to clustering and analyzing ehr
  opioid use trajectories for clinical subtypes.
\newblock {\em Journal of Biomedical Informatics}, 122:103889, 2021.

\bibitem{nguyen2016mathtt}
Phuoc Nguyen, Truyen Tran, Nilmini Wickramasinghe, and Svetha Venkatesh.
\newblock Deepr: a convolutional net for medical records.
\newblock {\em IEEE journal of biomedical and health informatics},
  21(1):22--30, 2016.

\bibitem{post2008clinical}
Bart Post, Johannes~D Speelman, and Rob~J De~Haan.
\newblock Clinical heterogeneity in newly diagnosed parkinson’s disease.
\newblock {\em Journal of neurology}, 255(5):716--722, 2008.

\bibitem{rabhi2022optimized}
Sara Rabhi.
\newblock {\em Optimized deep learning-based multimodal method for irregular
  medical timestamped data}.
\newblock PhD thesis, Institut Polytechnique de Paris, 2022.

\bibitem{rabhi2022temporal}
Sara Rabhi, Fr{\'e}d{\'e}ric Blanchard, Alpha~Mamadou Diallo, Djamal Zeghlache,
  C{\'e}line Lukas, Aur{\'e}lie Berot, Brigitte Delemer, and Sara Barraud.
\newblock Temporal deep learning framework for retinopathy prediction in
  patients with type 1 diabetes.
\newblock {\em Artificial Intelligence in Medicine}, page 102408, 2022.

\bibitem{ravanelli2017improving}
Mirco Ravanelli, Philemon Brakel, Maurizio Omologo, and Yoshua Bengio.
\newblock Improving speech recognition by revising gated recurrent units.
\newblock {\em arXiv preprint arXiv:1710.00641}, 2017.

\bibitem{rizopoulos2011dynamic}
Dimitris Rizopoulos.
\newblock Dynamic predictions and prospective accuracy in joint models for
  longitudinal and time-to-event data.
\newblock {\em Biometrics}, 67(3):819--829, 2011.

\bibitem{saito2015precision}
Takaya Saito and Marc Rehmsmeier.
\newblock The precision-recall plot is more informative than {ROC} plot when
  evaluating binary classifiers on imbalanced datasets.
\newblock {\em PloS One}, 10(3), 2015.

\bibitem{schulam2015framework}
Peter Schulam and Suchi Saria.
\newblock A framework for individualizing predictions of disease trajectories
  by exploiting multi-resolution structure.
\newblock In {\em Advances in NeurIPS}, pages 748--756, 2015.

\bibitem{severson20a}
Kristen~A. Severson, Lana~M. Chahine, Luba Smolensky, Kenney Ng, Jianying Hu,
  and Soumya Ghosh.
\newblock Personalized input-output hidden markov models for disease
  progression modeling.
\newblock In Finale Doshi-Velez, Jim Fackler, Ken Jung, David Kale, Rajesh
  Ranganath, Byron Wallace, and Jenna Wiens, editors, {\em Proceedings of the
  5th Machine Learning for Healthcare Conference}, volume 126 of {\em
  Proceedings of Machine Learning Research}, pages 309--330. PMLR, 07--08 Aug
  2020.

\bibitem{shalev2011online}
Shai Shalev-Shwartz et~al.
\newblock Online learning and online convex optimization.
\newblock 2011.

\bibitem{shamout2019deep}
Farah~E Shamout, Tingting Zhu, Pulkit Sharma, Peter~J Watkinson, and David~A
  Clifton.
\newblock Deep interpretable early warning system for the detection of clinical
  deterioration.
\newblock {\em IEEE journal of biomedical and health informatics},
  24(2):437--446, 2019.

\bibitem{sharafoddini2018finding}
Anis Sharafoddini, Joel~A Dubin, and Joon Lee.
\newblock Finding similar patient subpopulations in the icu using laboratory
  test ordering patterns.
\newblock In {\em Proceedings of the 2018 7th International Conference on
  Bioinformatics and Biomedical Science}, pages 72--77, 2018.

\bibitem{sharafoddini2019new}
Anis Sharafoddini, Joel~A Dubin, David~M Maslove, Joon Lee, et~al.
\newblock A new insight into missing data in intensive care unit patient
  profiles: observational study.
\newblock {\em JMIR medical informatics}, 7(1):e11605, 2019.

\bibitem{sutskever2014sequence}
I~Sutskever, O~Vinyals, and QV~Le.
\newblock Sequence to sequence learning with neural networks.
\newblock {\em Advances in NIPS}, pages 3104--12, 2014.

\bibitem{tan2020data}
Qingxiong Tan, Mang Ye, Baoyao Yang, Siqi Liu, Andy~Jinhua Ma, Terry Cheuk-Fung
  Yip, Grace Lai-Hung Wong, and PongChi Yuen.
\newblock Data-gru: Dual-attention time-aware gated recurrent unit for
  irregular multivariate time series.
\newblock In {\em Proceedings of the AAAI Conference on Artificial
  Intelligence}, volume~34, pages 930--937, 2020.

\bibitem{tran2015learning}
Truyen Tran, Tu~Dinh Nguyen, Dinh Phung, and Svetha Venkatesh.
\newblock Learning vector representation of medical objects via emr-driven
  nonnegative restricted boltzmann machines (enrbm).
\newblock {\em Journal of biomedical informatics}, 54:96--105, 2015.

\bibitem{tripoliti2017heart}
Evanthia~E Tripoliti, Theofilos~G Papadopoulos, Georgia~S Karanasiou,
  Katerina~K Naka, and Dimitrios~I Fotiadis.
\newblock Heart failure: diagnosis, severity estimation and prediction of
  adverse events through machine learning techniques.
\newblock {\em Computational and structural biotechnology journal}, 15:26--47,
  2017.

\bibitem{unnikrishnan2021love}
Vishnu Unnikrishnan, Yash Shah, Miro Schleicher, Carlos Fern{\'a}ndez-Viadero,
  Mirela Strandzheva, Doroteya Velikova, Plamen Dimitrov, Ruediger Pryss,
  Johannes Schobel, Winfried Schlee, et~al.
\newblock Love thy neighbours: a framework for error-driven discovery of useful
  neighbourhoods for one-step forecasts on ema data.
\newblock In {\em 2021 IEEE 34th International Symposium on Computer-Based
  Medical Systems (CBMS)}, pages 295--300. IEEE, 2021.

\bibitem{unnikrishnan2020predicting}
Vishnu Unnikrishnan, Yash Shah, Miro Schleicher, Mirela Strandzheva, Plamen
  Dimitrov, Doroteya Velikova, Ruediger Pryss, Johannes Schobel, Winfried
  Schlee, and Myra Spiliopoulou.
\newblock Predicting the health condition of mhealth app users with large
  differences in the number of recorded observations-where to learn from?
\newblock In {\em Discovery Science: 23rd International Conference, DS 2020,
  Thessaloniki, Greece, October 19--21, 2020, Proceedings 23}, pages 659--673.
  Springer, 2020.

\bibitem{van2010identification}
Stephanie~M Van~Rooden, Willem~J Heiser, Joost~N Kok, Dagmar Verbaan, Jacobus~J
  Van~Hilten, and Johan Marinus.
\newblock The identification of parkinson's disease subtypes using cluster
  analysis: a systematic review.
\newblock {\em Movement disorders}, 25(8):969--978, 2010.

\bibitem{visweswaran2005instance}
Shyam Visweswaran and Gregory~F Cooper.
\newblock Instance-specific bayesian model averaging for classification.
\newblock In {\em Advances in Neural Information Processing Systems}, pages
  1449--1456, 2005.

\bibitem{wicki2000predicting}
Jacques Wicki, Arnaud Perrier, Thomas~V Perneger, Henri Bounameaux, and
  Alain~Fran{\c{c}}ois Junod.
\newblock Predicting adverse outcome in patients with acute pulmonary embolism:
  a risk score.
\newblock {\em Thrombosis and haemostasis}, 84(10):548--552, 2000.

\bibitem{yu2020monitoring}
Ke~Yu, Mingda Zhang, Tianyi Cui, and Milos Hauskrecht.
\newblock Monitoring icu mortality risk with a long short-term memory recurrent
  neural network.
\newblock In {\em Pac Symp Biocomput}. World Scientific, 2020.

\bibitem{zhang2021interpretable}
Dongdong Zhang, Changchang Yin, Katherine~M Hunold, Xiaoqian Jiang, Jeffrey~M
  Caterino, and Ping Zhang.
\newblock An interpretable deep-learning model for early prediction of sepsis
  in the emergency department.
\newblock {\em Patterns}, 2(2):100196, 2021.

\bibitem{zhang2018patient2vec}
Jinghe Zhang, Kamran Kowsari, James~H Harrison, Jennifer~M Lobo, and Laura~E
  Barnes.
\newblock Patient2vec: A personalized interpretable deep representation of the
  longitudinal electronic health record.
\newblock {\em IEEE Access}, 6:65333--65346, 2018.

\bibitem{zhang2019data}
Xi~Zhang, Jingyuan Chou, Jian Liang, Cao Xiao, Yize Zhao, Harini Sarva, Claire
  Henchcliffe, and Fei Wang.
\newblock Data-driven subtyping of parkinson’s disease using longitudinal
  clinical records: a cohort study.
\newblock {\em Scientific reports}, 9(1):1--12, 2019.

\end{thebibliography}

\section*{Appendix}
\label{full-event-performance}
\scriptsize{
\begin{longtable}{@{}p{0.27\linewidth} | p{0.06\linewidth} | p{0.06\linewidth} | p{0.06\linewidth} | p{0.06\linewidth} | p{0.06\linewidth} | p{0.06\linewidth} | p{0.06\linewidth} | p{0.08\linewidth} | p{0.08\linewidth} | p{0.07\linewidth} @{}}
\caption{Performance of each lab test event, sorted by performance gain between Meta-Switch-Event vs. GRU-POP}\\
\toprule
\textbf{Lab Test Event Types}   & \textbf{Freq.}    & \textbf{CNN} & \textbf{RE TAIN}  & \textbf{GRU-POP} & \textbf{Sub-pop Adap} & \textbf{Self Adap} & \textbf{Com bined Adap}  & \textbf{Meta-Switch} & \textbf{Meta-Switch-Event} & \textbf{Gain (GRU-POP vs. Meta-Switch-Event.}\\ \midrule
{[}Chemistry/Blood{]} Benzodiazepine Screen & 0.002 & 0.17 & 0.50 & 0.50 & 0.45 & 0.28 & 14.40 & 15.12 & 24.10 & 6811.2 \\
{[}Chemistry/Urine{]} Benzodiazepine Screen, Urine & 0.002 & 0.23 & 2.40 & 1.38 & 0.86 & 0.31 & 34.91 & 24.82 & 29.56 & 5511.0 \\
{[}Chemistry/Urine{]} Opiate Screen, Urine & 0.002 & 0.19 & 0.99 & 0.95 & 0.53 & 0.33 & 11.27 & 11.69 & 16.68 & 5075.9 \\
{[}Chemistry/Blood{]} C-Reactive Protein & 0.006 & 0.62 & 1.11 & 1.73 & 0.91 & 1.13 & 7.97 & 8.31 & 7.68 & 2916.4 \\
{[}Chemistry/Urine{]} Methadone, Urine & 0.001 & 0.13 & 1.18 & 1.18 & 0.62 & 0.37 & 1.76 & 1.21 & 1.23 & 2548.7 \\
{[}Chemistry/Urine{]} Barbiturate Screen, Urine & 0.001 & 0.12 & 0.89 & 1.19 & 0.70 & 0.38 & 2.28 & 1.17 & 1.47 & 2365.8 \\
{[}Chemistry/Urine{]} Amphetamine Screen, Urine & 0.001 & 0.12 & 0.93 & 1.22 & 0.70 & 0.38 & 2.27 & 1.30 & 1.46 & 2345.4 \\
{[}Chemistry/Urine{]} Cocaine, Urine & 0.001 & 0.13 & 1.24 & 1.34 & 0.68 & 0.37 & 1.82 & 1.20 & 1.31 & 2263.6 \\
{[}Chemistry/Urine{]} Protein/Creatinine Ratio & 0.005 & 0.50 & 1.09 & 1.80 & 1.44 & 1.32 & 3.69 & 8.78 & 8.11 & 1689.6 \\
{[}Chemistry/Blood{]} Thyroxine (T4), Free & 0.005 & 0.53 & 0.58 & 0.71 & 0.88 & 0.73 & 2.47 & 1.92 & 2.81 & 1512.3 \\
{[}Chemistry/Blood{]} Digoxin & 0.013 & 1.30 & 2.97 & 5.27 & 3.33 & 2.04 & 25.87 & 24.68 & 25.51 & 1496.9 \\
{[}Hematology/Blood{]} Sedimentation Rate & 0.005 & 0.52 & 1.42 & 1.85 & 1.58 & 1.14 & 6.67 & 6.89 & 6.31 & 1388.8 \\
{[}Chemistry/Urine{]} Total Protein, Urine & 0.006 & 0.56 & 1.32 & 2.45 & 2.06 & 1.49 & 3.77 & 8.23 & 8.05 & 1106.2 \\
{[}Blood Gas/Blood{]} Calculated Bicarbonate, Whole Blood & 0.005 & 0.54 & 2.51 & 1.43 & 1.39 & 0.86 & 2.93 & 2.64 & 3.46 & 973.5 \\
{[}Chemistry/Blood{]} Cholesterol, LDL, Calculated & 0.004 & 0.38 & 0.95 & 0.63 & 0.80 & 0.63 & 0.63 & 0.53 & 0.75 & 971.9 \\
{[}Chemistry/Blood{]} Cholesterol, Total & 0.005 & 0.54 & 1.01 & 0.78 & 0.95 & 0.77 & 0.85 & 0.73 & 0.91 & 920.8 \\
{[}Chemistry/Blood{]} Cholesterol, HDL & 0.004 & 0.44 & 0.94 & 0.69 & 0.89 & 0.72 & 0.76 & 0.67 & 0.83 & 912.9 \\
{[}Chemistry/Blood{]} Cholesterol Ratio (Total/HDL) & 0.004 & 0.41 & 0.95 & 0.68 & 0.89 & 0.73 & 0.75 & 0.69 & 0.85 & 909.9 \\
{[}Chemistry/Blood{]} Vitamin B12 & 0.006 & 0.58 & 1.78 & 1.16 & 1.16 & 0.86 & 0.68 & 0.69 & 0.84 & 872.5 \\
{[}Chemistry/Blood{]} Ferritin & 0.013 & 1.34 & 1.78 & 1.75 & 1.68 & 1.43 & 2.49 & 1.78 & 2.03 & 789.7 \\
{[}Chemistry/Blood{]} Folate & 0.005 & 0.50 & 3.46 & 0.91 & 1.29 & 0.63 & 0.71 & 0.55 & 0.78 & 785.6 \\
{[}Hematology/Urine{]} Transitional Epithelial Cells & 0.007 & 0.71 & 1.14 & 1.20 & 1.35 & 0.87 & 0.89 & 1.06 & 1.48 & 754.1 \\
{[}Chemistry/Blood{]} Protein, Total & 0.006 & 0.57 & 0.83 & 1.53 & 1.39 & 1.03 & 1.68 & 1.19 & 1.47 & 738.7 \\
{[}Chemistry/Blood{]} Transferrin & 0.014 & 1.42 & 1.72 & 2.57 & 1.80 & 1.63 & 1.41 & 1.52 & 1.84 & 677.0 \\
{[}Chemistry/Blood{]} Iron Binding Capacity, Total & 0.014 & 1.42 & 1.72 & 2.57 & 1.80 & 1.63 & 1.41 & 1.52 & 1.84 & 673.5 \\
{[}Chemistry/Blood{]} Iron & 0.015 & 1.46 & 1.80 & 2.59 & 1.87 & 1.65 & 1.42 & 1.54 & 1.83 & 662.1 \\
{[}Chemistry/Blood{]} NTproBNP & 0.006 & 0.56 & 1.44 & 1.75 & 1.68 & 0.71 & 1.77 & 1.13 & 1.93 & 643.0 \\
{[}Chemistry/Blood{]} Barbiturate Screen & 0.001 & 0.08 & 0.95 & 0.58 & 0.80 & 0.33 & 0.69 & 0.23 & 0.80 & 589.9 \\
{[}Chemistry/Blood{]} Tricyclic Antidepressant Screen & 0.001 & 0.09 & 0.90 & 0.46 & 0.82 & 0.31 & 0.62 & 0.21 & 0.75 & 577.1 \\
{[}Chemistry/Blood{]} Thyroid Stimulating Hormone & 0.018 & 1.77 & 2.48 & 2.27 & 2.32 & 2.11 & 2.22 & 2.71 & 2.76 & 576.9 \\
{[}Hematology/Blood{]} Reticulocyte Count, Automated & 0.013 & 1.28 & 2.20 & 2.86 & 2.74 & 1.91 & 2.17 & 1.81 & 2.44 & 557.2 \\
{[}Chemistry/Blood{]} Red Top Hold & 0.004 & 0.44 & 1.67 & 2.62 & 2.36 & 1.77 & 2.06 & 1.58 & 2.07 & 552.5 \\
{[}Hematology/Other Body Fluid{]} Monos & 0.006 & 0.58 & 1.23 & 2.08 & 2.03 & 1.57 & 1.65 & 1.24 & 3.77 & 549.7 \\
{[}Hematology/Other Body Fluid{]} Polys & 0.006 & 0.58 & 1.21 & 2.08 & 2.03 & 1.57 & 1.65 & 1.24 & 3.76 & 549.5 \\
{[}Hematology/Other Body Fluid{]} Lymphocytes & 0.006 & 0.58 & 1.25 & 2.08 & 2.04 & 1.57 & 1.65 & 1.24 & 3.77 & 549.0 \\
{[}Hematology/Urine{]} Granular Casts & 0.016 & 1.56 & 3.44 & 3.25 & 2.80 & 1.83 & 2.39 & 2.22 & 2.94 & 494.5 \\
{[}Hematology/Urine{]} Amorphous Crystals & 0.010 & 0.98 & 1.89 & 2.28 & 2.43 & 1.44 & 1.96 & 1.94 & 2.50 & 476.5 \\
{[}Blood Gas/Blood{]} Alveolar-arterial Gradient & 0.017 & 1.71 & 3.41 & 6.13 & 5.17 & 3.37 & 12.65 & 9.11 & 13.15 & 456.3 \\
{[}Blood Gas/Blood{]} Required O2 & 0.017 & 1.71 & 3.51 & 6.18 & 5.21 & 3.38 & 12.68 & 9.11 & 13.17 & 452.2 \\
{[}Chemistry/Blood{]} Triglycerides & 0.023 & 2.26 & 5.06 & 6.44 & 8.21 & 3.81 & 11.67 & 9.53 & 8.89 & 350.4 \\
{[}Chemistry/Blood{]} Blue Top Hold & 0.009 & 0.86 & 3.90 & 3.81 & 4.12 & 2.33 & 3.66 & 4.18 & 5.72 & 347.9 \\
{[}Chemistry/Blood{]} \% Hemoglobin A1c & 0.003 & 0.32 & 1.03 & 0.94 & 2.11 & 1.16 & 0.76 & 0.76 & 1.03 & 330.7 \\
{[}Hematology/Urine{]} Urine Mucous & 0.018 & 1.81 & 6.50 & 5.70 & 4.90 & 3.43 & 3.61 & 3.50 & 6.99 & 307.3 \\
{[}Blood Gas/Blood{]} O2 Flow & 0.017 & 1.73 & 5.70 & 6.13 & 6.62 & 3.26 & 5.07 & 6.47 & 6.64 & 306.5 \\
{[}Chemistry/Blood{]} Cortisol & 0.013 & 1.27 & 3.30 & 4.49 & 5.23 & 2.46 & 3.25 & 4.02 & 3.67 & 292.0 \\
{[}Hematology/Urine{]} Hyaline Casts & 0.036 & 3.59 & 5.54 & 6.25 & 5.79 & 4.16 & 4.90 & 4.63 & 6.80 & 289.3 \\
{[}Hematology/Blood{]} Fibrin Degradation Products & 0.012 & 1.23 & 12.67 & 12.18 & 12.95 & 5.02 & 20.96 & 22.96 & 24.73 & 287.7 \\
{[}Chemistry/Urine{]} Potassium, Urine & 0.033 & 3.34 & 8.98 & 8.15 & 7.45 & 5.27 & 6.67 & 6.86 & 9.55 & 270.1 \\
{[}Chemistry/Urine{]} Chloride, Urine & 0.031 & 3.11 & 7.87 & 8.44 & 7.16 & 4.79 & 5.91 & 5.86 & 9.97 & 266.7 \\
{[}Hematology/Blood{]} Nucleated Red Cells & 0.017 & 1.68 & 12.61 & 14.40 & 12.54 & 6.65 & 19.19 & 17.25 & 20.69 & 262.4 \\
{[}Chemistry/Blood{]} Haptoglobin & 0.022 & 2.21 & 6.26 & 8.33 & 7.45 & 5.07 & 6.50 & 6.79 & 8.05 & 260.8 \\
{[}Chemistry/Blood{]} Bilirubin, Direct & 0.031 & 3.14 & 11.55 & 15.72 & 13.61 & 6.64 & 17.14 & 15.31 & 17.62 & 244.9 \\
{[}Chemistry/Blood{]} Bilirubin, Indirect & 0.030 & 2.95 & 11.67 & 15.57 & 13.59 & 6.73 & 17.09 & 15.45 & 17.69 & 244.5 \\
{[}Chemistry/Urine{]} Urea Nitrogen, Urine & 0.030 & 2.97 & 7.38 & 7.07 & 7.55 & 5.12 & 6.01 & 5.41 & 7.43 & 225.5 \\
{[}Chemistry/Urine{]} Osmolality, Urine & 0.037 & 3.65 & 8.30 & 10.19 & 8.50 & 5.46 & 7.14 & 6.93 & 9.64 & 223.4 \\
{[}Blood Gas/Blood{]} Chloride, Whole Blood & 0.033 & 3.27 & 12.63 & 11.87 & 12.50 & 7.97 & 12.20 & 12.18 & 18.31 & 210.6 \\
{[}Chemistry/Urine{]} Sodium, Urine & 0.046 & 4.62 & 9.18 & 11.20 & 9.45 & 6.52 & 7.77 & 7.18 & 10.77 & 210.0 \\
{[}Chemistry/Urine{]} Length of Urine Collection & 0.023 & 2.32 & 11.08 & 13.58 & 11.25 & 6.73 & 11.88 & 10.96 & 14.12 & 209.0 \\
{[}Hematology/Urine{]} Bacteria & 0.059 & 5.86 & 9.49 & 11.49 & 11.14 & 8.44 & 9.65 & 9.40 & 16.82 & 208.0 \\
{[}Chemistry/Urine{]} Creatinine, Urine & 0.051 & 5.06 & 10.40 & 11.68 & 10.33 & 7.35 & 8.94 & 8.90 & 11.17 & 203.7 \\
{[}Hematology/Urine{]} RBC & 0.096 & 9.58 & 11.31 & 13.45 & 12.98 & 10.91 & 11.40 & 11.51 & 17.46 & 195.0 \\
{[}Blood Gas/Blood{]} Sodium, Whole Blood & 0.041 & 4.10 & 15.09 & 14.41 & 14.44 & 9.97 & 14.20 & 14.11 & 21.21 & 194.9 \\
{[}Hematology/Urine{]} WBC & 0.099 & 9.88 & 11.67 & 13.28 & 12.89 & 11.33 & 11.20 & 11.22 & 16.56 & 193.6 \\
{[}Hematology/Urine{]} Yeast & 0.102 & 10.23 & 11.91 & 13.75 & 13.25 & 11.69 & 11.67 & 11.80 & 17.03 & 189.8 \\
{[}Hematology/Urine{]} Epithelial Cells & 0.095 & 9.49 & 11.14 & 13.21 & 12.63 & 10.89 & 10.83 & 10.65 & 16.22 & 189.4 \\
{[}Hematology/Urine{]} Protein & 0.104 & 10.44 & 12.60 & 15.57 & 15.04 & 12.64 & 13.26 & 13.16 & 20.04 & 176.2 \\
{[}Hematology/Urine{]} Glucose & 0.079 & 7.86 & 11.69 & 14.56 & 14.51 & 11.42 & 12.56 & 12.37 & 20.06 & 170.8 \\
{[}Hematology/Urine{]} Urine Color & 0.079 & 7.88 & 11.67 & 14.37 & 14.47 & 11.66 & 12.51 & 12.31 & 20.27 & 167.6 \\
{[}Hematology/Urine{]} Urobilinogen & 0.083 & 8.29 & 12.07 & 14.62 & 14.54 & 11.53 & 12.39 & 12.16 & 19.87 & 166.7 \\
{[}Hematology/Urine{]} Ketone & 0.081 & 8.08 & 11.88 & 14.77 & 14.54 & 11.54 & 12.37 & 12.11 & 19.74 & 166.3 \\
{[}Hematology/Urine{]} Leukocytes & 0.075 & 7.55 & 11.77 & 14.54 & 14.56 & 11.50 & 12.55 & 12.37 & 20.35 & 165.7 \\
{[}Hematology/Urine{]} Nitrite & 0.075 & 7.55 & 11.77 & 14.53 & 14.57 & 11.50 & 12.56 & 12.35 & 20.33 & 165.3 \\
{[}Hematology/Urine{]} Blood & 0.076 & 7.56 & 11.77 & 14.51 & 14.58 & 11.50 & 12.54 & 12.33 & 20.34 & 164.9 \\
{[}Hematology/Urine{]} Bilirubin & 0.075 & 7.55 & 11.75 & 14.52 & 14.59 & 11.50 & 12.55 & 12.37 & 20.32 & 164.9 \\
{[}Hematology/Urine{]} Urine Appearance & 0.076 & 7.59 & 11.76 & 14.53 & 14.58 & 11.51 & 12.47 & 12.30 & 20.21 & 164.0 \\
{[}Hematology/Urine{]} pH & 0.128 & 12.83 & 14.79 & 17.34 & 16.94 & 14.52 & 14.84 & 14.79 & 20.70 & 163.8 \\
{[}Hematology/Urine{]} Specific Gravity & 0.128 & 12.77 & 14.73 & 17.28 & 16.91 & 14.47 & 14.68 & 14.69 & 20.51 & 162.7 \\
{[}Chemistry/Blood{]} Lipase & 0.063 & 6.31 & 22.39 & 26.48 & 22.95 & 15.84 & 28.30 & 26.65 & 29.92 & 161.2 \\
{[}Blood Gas/Blood{]} Hematocrit, Calculated & 0.036 & 3.63 & 15.31 & 14.00 & 14.84 & 9.37 & 13.62 & 13.27 & 17.66 & 156.4 \\
{[}Blood Gas/Blood{]} Hemoglobin & 0.036 & 3.63 & 15.46 & 14.01 & 14.83 & 9.36 & 13.59 & 13.26 & 17.65 & 156.4 \\
{[}Hematology/Blood{]} Ovalocytes & 0.008 & 0.82 & 8.57 & 12.80 & 12.61 & 7.12 & 8.41 & 8.41 & 10.53 & 151.9 \\
{[}Hematology/Blood{]} Platelet Smear & 0.020 & 2.00 & 19.11 & 21.05 & 21.32 & 13.14 & 23.75 & 23.07 & 28.03 & 130.9 \\
{[}Chemistry/Blood{]} Uric Acid & 0.013 & 1.34 & 29.28 & 55.57 & 34.24 & 14.25 & 58.65 & 59.16 & 58.09 & 126.4 \\
{[}Chemistry/Blood{]} Amylase & 0.055 & 5.55 & 25.31 & 29.03 & 26.56 & 17.02 & 27.90 & 27.69 & 30.53 & 121.3 \\
{[}Chemistry/Blood{]} Troponin T & 0.055 & 5.45 & 23.30 & 25.24 & 26.39 & 18.14 & 26.29 & 25.99 & 30.05 & 115.5 \\
{[}Chemistry/Blood{]} Creatine Kinase (CK) & 0.086 & 8.64 & 30.22 & 31.47 & 30.75 & 23.48 & 30.89 & 29.64 & 33.11 & 103.3 \\
{[}Chemistry/Blood{]} Creatine Kinase, MB Isoenzyme & 0.063 & 6.34 & 27.39 & 27.30 & 27.06 & 20.59 & 24.27 & 24.13 & 27.42 & 102.6 \\
{[}Hematology/Blood{]} Polychromasia & 0.020 & 1.96 & 20.66 & 24.49 & 24.19 & 14.41 & 20.91 & 20.57 & 26.58 & 100.7 \\
{[}Hematology/Blood{]} Macrocytes & 0.020 & 1.96 & 20.69 & 24.52 & 24.20 & 14.41 & 20.91 & 20.57 & 26.58 & 100.7 \\
{[}Hematology/Blood{]} Microcytes & 0.020 & 1.96 & 20.65 & 24.52 & 24.19 & 14.40 & 20.89 & 20.57 & 26.59 & 100.7 \\
{[}Hematology/Blood{]} Poikilocytosis & 0.020 & 1.96 & 20.68 & 24.51 & 24.20 & 14.40 & 20.89 & 20.57 & 26.59 & 100.7 \\
{[}Hematology/Blood{]} Hypochromia & 0.020 & 1.96 & 20.70 & 24.50 & 24.20 & 14.40 & 20.89 & 20.58 & 26.59 & 100.7 \\
{[}Hematology/Blood{]} Anisocytosis & 0.020 & 1.96 & 20.65 & 24.49 & 24.20 & 14.40 & 20.89 & 20.57 & 26.59 & 100.6 \\
{[}Hematology/Blood{]} Atypical Lymphocytes & 0.061 & 6.08 & 32.85 & 34.66 & 32.45 & 20.98 & 33.31 & 30.90 & 37.74 & 86.4 \\
{[}Hematology/Blood{]} Myelocytes & 0.061 & 6.09 & 32.91 & 34.69 & 32.50 & 21.01 & 33.28 & 30.85 & 37.70 & 85.8 \\
{[}Hematology/Blood{]} Metamyelocytes & 0.061 & 6.09 & 32.92 & 34.70 & 32.52 & 21.01 & 33.28 & 30.84 & 37.70 & 85.7 \\
{[}Blood Gas/Blood{]} Ventilator & 0.035 & 3.48 & 30.04 & 30.07 & 32.41 & 24.09 & 28.65 & 29.70 & 31.56 & 82.2 \\
{[}Hematology/Blood{]} Eosinophils & 0.140 & 14.05 & 35.07 & 37.42 & 35.39 & 25.51 & 33.85 & 32.87 & 38.92 & 78.1 \\
{[}Hematology/Blood{]} Basophils & 0.140 & 14.05 & 35.07 & 37.42 & 35.39 & 25.51 & 33.84 & 32.86 & 38.89 & 78.1 \\
{[}Hematology/Blood{]} Monocytes & 0.140 & 14.05 & 35.07 & 37.44 & 35.39 & 25.51 & 33.85 & 32.86 & 38.91 & 78.1 \\
{[}Hematology/Blood{]} Lymphocytes & 0.140 & 14.05 & 35.11 & 37.39 & 35.40 & 25.52 & 33.86 & 32.87 & 38.92 & 78.0 \\
{[}Hematology/Blood{]} Neutrophils & 0.140 & 14.05 & 35.09 & 37.42 & 35.42 & 25.52 & 33.86 & 32.87 & 38.93 & 78.0 \\
{[}Hematology/Blood{]} Bands & 0.066 & 6.63 & 33.84 & 35.39 & 33.82 & 21.92 & 32.93 & 30.68 & 37.85 & 76.3 \\
{[}Chemistry/Blood{]} CK-MB Index & 0.012 & 1.24 & 35.22 & 30.10 & 29.95 & 18.79 & 24.33 & 23.17 & 25.69 & 75.4 \\
{[}Chemistry/Blood{]} Vancomycin & 0.163 & 16.28 & 37.69 & 41.56 & 43.60 & 35.45 & 43.17 & 42.28 & 44.96 & 69.7 \\
{[}Chemistry/Blood{]} Lactate Dehydrogenase (LD) & 0.176 & 17.55 & 43.05 & 46.33 & 43.14 & 32.72 & 44.28 & 42.45 & 47.22 & 69.4 \\
{[}Chemistry/Blood{]} Albumin & 0.174 & 17.36 & 40.20 & 42.62 & 40.26 & 31.57 & 37.46 & 37.54 & 41.57 & 63.4 \\
{[}Hematology/Blood{]} Fibrinogen, Functional & 0.070 & 7.02 & 40.69 & 47.27 & 43.11 & 31.26 & 44.59 & 42.20 & 46.07 & 62.5 \\
{[}Chemistry/Blood{]} Osmolality, Measured & 0.027 & 2.67 & 35.22 & 45.82 & 45.28 & 31.38 & 47.90 & 46.69 & 48.85 & 59.1 \\
{[}Chemistry/Blood{]} Estimated GFR (MDRD equation) & 0.028 & 2.81 & 13.95 & 9.92 & 12.11 & 9.20 & 4.53 & 6.80 & 6.16 & 58.3 \\
{[}Blood Gas/Blood{]} Tidal Volume & 0.113 & 11.25 & 45.95 & 47.65 & 46.82 & 39.15 & 44.31 & 45.73 & 48.57 & 57.1 \\
{[}Chemistry/Blood{]} Phenytoin & 0.045 & 4.47 & 49.15 & 60.11 & 56.67 & 48.80 & 60.53 & 59.29 & 59.60 & 49.3 \\
{[}Blood Gas/Blood{]} Ventilation Rate & 0.048 & 4.85 & 49.27 & 48.08 & 52.19 & 45.79 & 46.03 & 48.73 & 51.37 & 41.7 \\
{[}Blood Gas/Blood{]} PEEP & 0.144 & 14.39 & 54.64 & 56.14 & 55.38 & 47.14 & 52.93 & 52.72 & 57.01 & 41.2 \\
{[}Blood Gas/Blood{]} Oxygen & 0.162 & 16.24 & 55.16 & 56.87 & 55.57 & 46.86 & 52.27 & 52.01 & 56.00 & 40.2 \\
{[}Blood Gas/Blood{]} Oxygen Saturation & 0.180 & 18.04 & 58.43 & 61.64 & 60.46 & 52.75 & 59.39 & 60.22 & 63.33 & 36.2 \\
{[}Blood Gas/Blood{]} Intubated & 0.081 & 8.13 & 58.46 & 58.00 & 60.95 & 53.53 & 57.61 & 56.73 & 61.59 & 33.7 \\
{[}Blood Gas/Blood{]} Lactate & 0.274 & 27.36 & 62.73 & 62.79 & 63.67 & 56.42 & 57.73 & 60.67 & 63.37 & 33.2 \\
{[}Blood Gas/Blood{]} Temperature & 0.212 & 21.18 & 61.49 & 62.06 & 61.91 & 54.31 & 58.65 & 58.74 & 62.87 & 33.1 \\
{[}Blood Gas/Blood{]} Glucose & 0.157 & 15.66 & 62.29 & 63.94 & 63.87 & 56.17 & 59.67 & 61.25 & 64.89 & 28.9 \\
{[}Blood Gas/Blood{]} Potassium, Whole Blood & 0.147 & 14.69 & 61.37 & 63.88 & 63.54 & 55.54 & 58.53 & 60.19 & 64.36 & 28.8 \\
{[}Chemistry/Blood{]} Alkaline Phosphatase & 0.277 & 27.71 & 66.94 & 69.28 & 68.55 & 59.41 & 67.12 & 66.30 & 70.94 & 26.5 \\
{[}Chemistry/Blood{]} Bilirubin, Total & 0.285 & 28.54 & 67.43 & 69.68 & 68.77 & 59.44 & 67.62 & 66.29 & 71.11 & 26.1 \\
{[}Chemistry/Blood{]} Alanine Aminotransferase (ALT) & 0.279 & 27.95 & 67.44 & 69.90 & 69.11 & 59.62 & 67.64 & 66.49 & 71.41 & 25.7 \\
{[}Chemistry/Blood{]} Asparate Aminotransferase (AST) & 0.280 & 27.96 & 67.35 & 69.78 & 69.04 & 59.70 & 67.63 & 66.45 & 71.33 & 25.7 \\
{[}Blood Gas/Blood{]} Free Calcium & 0.310 & 31.00 & 73.95 & 75.51 & 75.80 & 69.65 & 72.19 & 73.48 & 76.05 & 18.8 \\
{[}Hematology/Blood{]} PT & 0.555 & 55.51 & 80.33 & 81.85 & 80.56 & 74.48 & 78.77 & 78.60 & 81.35 & 16.7 \\
{[}Hematology/Blood{]} INR(PT) & 0.555 & 55.51 & 80.29 & 81.85 & 80.56 & 74.48 & 78.77 & 78.60 & 81.35 & 16.7 \\
{[}Hematology/Blood{]} PTT & 0.547 & 54.71 & 80.71 & 82.42 & 80.90 & 74.97 & 78.92 & 78.83 & 81.53 & 16.1 \\
{[}BLOOD GAS/BLOOD{]} SPECIMEN TYPE & 0.163 & 16.32 & 83.80 & 82.27 & 85.71 & 80.29 & 81.17 & 83.07 & 84.62 & 11.0 \\
{[}Blood Gas/Blood{]} pO2 & 0.463 & 46.32 & 85.04 & 85.54 & 85.83 & 82.21 & 81.21 & 84.35 & 86.42 & 10.8 \\
{[}Blood Gas/Blood{]} pCO2 & 0.463 & 46.31 & 85.13 & 85.55 & 85.83 & 82.22 & 81.21 & 84.36 & 86.43 & 10.8 \\
{[}Blood Gas/Blood{]} Base Excess & 0.463 & 46.31 & 85.08 & 85.55 & 85.84 & 82.22 & 81.21 & 84.36 & 86.42 & 10.8 \\
{[}Blood Gas/Blood{]} Calculated Total CO2 & 0.463 & 46.33 & 85.14 & 85.60 & 85.91 & 82.25 & 81.25 & 84.41 & 86.50 & 10.7 \\
{[}Blood Gas/Blood{]} pH & 0.492 & 49.23 & 86.02 & 86.82 & 86.76 & 83.39 & 82.04 & 85.38 & 87.14 & 10.0 \\
{[}Chemistry/Blood{]} Calcium, Total & 0.874 & 87.42 & 93.27 & 93.23 & 93.93 & 92.79 & 91.34 & 93.66 & 95.40 & 5.6 \\
{[}Chemistry/Blood{]} Phosphate & 0.878 & 87.82 & 93.67 & 93.69 & 94.35 & 93.17 & 91.59 & 93.89 & 95.71 & 5.1 \\
{[}Hematology/Blood{]} MCV & 0.903 & 90.34 & 95.69 & 95.51 & 95.98 & 95.02 & 92.50 & 95.39 & 97.20 & 3.8 \\
{[}Hematology/Blood{]} MCH & 0.903 & 90.34 & 95.63 & 95.51 & 95.98 & 95.02 & 92.51 & 95.39 & 97.19 & 3.8 \\
{[}Hematology/Blood{]} Red Blood Cells & 0.903 & 90.34 & 95.65 & 95.51 & 95.99 & 95.01 & 92.50 & 95.39 & 97.19 & 3.8 \\
{[}Hematology/Blood{]} MCHC & 0.903 & 90.34 & 95.69 & 95.50 & 95.99 & 95.02 & 92.50 & 95.39 & 97.19 & 3.8 \\
{[}Hematology/Blood{]} White Blood Cells & 0.904 & 90.38 & 95.71 & 95.51 & 96.00 & 95.03 & 92.56 & 95.39 & 97.18 & 3.8 \\
{[}Hematology/Blood{]} RDW & 0.903 & 90.34 & 95.68 & 95.50 & 96.00 & 95.00 & 92.51 & 95.39 & 97.19 & 3.8 \\
{[}Hematology/Blood{]} Platelet Count & 0.904 & 90.39 & 95.74 & 95.55 & 96.02 & 95.13 & 92.50 & 95.48 & 97.11 & 3.8 \\
{[}Hematology/Blood{]} Hemoglobin & 0.904 & 90.36 & 95.74 & 95.55 & 96.03 & 95.05 & 92.55 & 95.42 & 97.19 & 3.8 \\
{[}Chemistry/Blood{]} Magnesium & 0.905 & 90.55 & 95.29 & 95.32 & 96.02 & 95.11 & 92.58 & 95.53 & 96.84 & 3.8 \\
{[}Hematology/Blood{]} Hematocrit & 0.908 & 90.84 & 96.21 & 96.04 & 96.51 & 95.60 & 93.08 & 95.92 & 97.46 & 3.3 \\
{[}Chemistry/Blood{]} Glucose & 0.909 & 90.91 & 96.13 & 95.72 & 96.50 & 95.78 & 93.06 & 96.12 & 97.71 & 3.3 \\
{[}Chemistry/Blood{]} Anion Gap & 0.908 & 90.80 & 96.17 & 95.77 & 96.53 & 95.77 & 93.17 & 96.21 & 97.82 & 3.3 \\
{[}Chemistry/Blood{]} Bicarbonate & 0.910 & 91.04 & 96.42 & 95.95 & 96.66 & 95.84 & 93.34 & 96.25 & 97.85 & 3.2 \\
{[}Chemistry/Blood{]} Creatinine & 0.912 & 91.20 & 96.56 & 95.99 & 96.76 & 95.86 & 93.31 & 96.25 & 97.81 & 3.1 \\
{[}Chemistry/Blood{]} Urea Nitrogen & 0.912 & 91.20 & 96.51 & 96.03 & 96.76 & 95.92 & 93.36 & 96.31 & 97.85 & 3.1 \\
{[}Chemistry/Blood{]} Sodium & 0.913 & 91.33 & 96.62 & 96.13 & 96.80 & 96.04 & 93.49 & 96.45 & 97.87 & 3.1 \\
{[}Chemistry/Blood{]} Chloride & 0.913 & 91.29 & 96.60 & 96.15 & 96.85 & 96.03 & 93.50 & 96.43 & 97.85 & 3.0 \\
{[}Chemistry/Blood{]} Potassium & 0.914 & 91.38 & 96.60 & 96.16 & 96.92 & 96.18 & 93.52 & 96.57 & 97.98 & 3.0 \\
Multi Lumen & 0.435 & 43.46 & 95.31 & 95.60 & 97.17 & 95.99 & 96.28 & 96.97 & 97.01 & 2.3 \\
\bottomrule
\end{longtable}
}

\scriptsize{
\begin{longtable}{@{}p{0.27\linewidth} | p{0.06\linewidth} | p{0.06\linewidth} | p{0.06\linewidth} | p{0.06\linewidth} | p{0.06\linewidth} | p{0.06\linewidth} | p{0.06\linewidth} | p{0.08\linewidth} | p{0.08\linewidth} | p{0.07\linewidth} @{}}
\caption{Performance of each medication administration event, sorted by performance gain between Meta-Switch-Event vs. GRU-POP}\\
\toprule
\textbf{Medication Administration Event Types}   & \textbf{Freq.}    & \textbf{CNN} & \textbf{RE TAIN}  & \textbf{GRU-POP} & \textbf{Sub-pop Adap} & \textbf{Self Adap} & \textbf{Com bined Adap}  & \textbf{Meta-Switch} & \textbf{Meta-Switch-Event} & \textbf{Gain (GRU-POP vs. Meta-Switch-Event.}\\ \midrule
{[}Medications{]} Insulin & 0.004 & 0.36 & 2.62 & 23.57 & 1.37 & 0.66 & 52.45 & 60.84 & 51.83 & 5933.8 \\
{[}Medications{]} Enoxaparin (Lovenox) & 0.015 & 1.46 & 6.75 & 34.31 & 7.50 & 5.83 & 47.34 & 41.17 & 44.77 & 877.3 \\
{[}Medications{]} Omeprazole (Prilosec) & 0.011 & 1.09 & 6.77 & 17.64 & 6.13 & 2.78 & 17.72 & 17.03 & 20.73 & 619.6 \\
{[}Medications{]} Lansoprazole (Prevacid) & 0.029 & 2.93 & 9.01 & 22.56 & 10.70 & 5.35 & 25.73 & 22.89 & 25.18 & 422.7 \\
{[}Medications{]} Na Phos & 0.019 & 1.89 & 6.34 & 8.49 & 11.20 & 4.91 & 23.64 & 21.52 & 24.60 & 366.8 \\
{[}Medications{]} Haloperidol (Haldol) & 0.033 & 3.31 & 7.10 & 12.06 & 10.01 & 5.89 & 16.94 & 16.21 & 18.62 & 353.3 \\
{[}Medications{]} Coumadin (Warfarin) & 0.015 & 1.50 & 7.14 & 12.43 & 9.60 & 6.46 & 12.16 & 11.70 & 12.55 & 345.9 \\
{[}Medications{]} Sodium Bicarbonate 8.4\% & 0.026 & 2.62 & 9.34 & 13.69 & 11.85 & 5.71 & 24.92 & 19.58 & 23.78 & 336.9 \\
{[}Antibiotics{]} Ampicillin & 0.015 & 1.54 & 23.38 & 48.21 & 21.15 & 10.07 & 60.51 & 57.38 & 63.37 & 289.0 \\
{[}Medications{]} Ranitidine (Prophylaxis) & 0.027 & 2.69 & 13.58 & 18.52 & 14.24 & 11.95 & 21.87 & 22.56 & 23.48 & 246.6 \\
{[}Antibiotics{]} Fluconazole & 0.025 & 2.50 & 21.47 & 47.83 & 24.57 & 11.85 & 59.32 & 59.27 & 58.81 & 234.2 \\
{[}Blood Products/Colloids{]} Fresh Frozen Plasma & 0.024 & 2.41 & 11.80 & 17.59 & 14.53 & 8.55 & 18.91 & 16.96 & 19.17 & 228.4 \\
{[}Antibiotics{]} Piperacillin & 0.032 & 3.19 & 12.74 & 27.93 & 20.28 & 9.63 & 31.38 & 32.10 & 32.87 & 211.7 \\
{[}Medications{]} K Phos & 0.067 & 6.69 & 18.00 & 25.17 & 21.08 & 16.60 & 32.14 & 30.64 & 32.39 & 206.1 \\
{[}Medications{]} Diltiazem & 0.024 & 2.44 & 13.82 & 27.76 & 21.00 & 9.56 & 32.45 & 31.07 & 32.18 & 202.5 \\
{[}Blood Products/Colloids{]} Albumin & 0.048 & 4.81 & 17.84 & 26.85 & 21.47 & 12.60 & 34.30 & 30.17 & 31.71 & 185.1 \\
{[}Medications{]} Dexmedetomidine (Precedex) & 0.030 & 3.03 & 14.53 & 29.22 & 24.64 & 12.83 & 39.66 & 38.42 & 40.63 & 180.4 \\
{[}Medications{]} Labetalol & 0.025 & 2.50 & 18.22 & 36.50 & 24.93 & 15.16 & 41.77 & 41.25 & 43.73 & 175.3 \\
{[}Antibiotics{]} Azithromycin & 0.014 & 1.43 & 23.02 & 44.58 & 27.32 & 18.32 & 50.71 & 43.43 & 42.83 & 164.9 \\
{[}Blood Products/Colloids{]} Albumin & 0.040 & 4.00 & 18.56 & 21.08 & 21.02 & 13.71 & 25.14 & 23.02 & 25.93 & 148.8 \\
{[}Medications{]} Amiodarone & 0.035 & 3.55 & 24.51 & 34.09 & 31.06 & 21.51 & 41.26 & 42.35 & 44.49 & 128.5 \\
{[}Blood Products/Colloids{]} OR FFP Intake & 0.003 & 0.34 & 6.26 & 5.73 & 4.38 & 5.35 & 1.22 & 2.05 & 2.11 & 122.6 \\
{[}Medications{]} Dopamine & 0.017 & 1.65 & 41.28 & 61.22 & 37.79 & 22.10 & 61.95 & 57.72 & 59.00 & 119.7 \\
{[}Antibiotics{]} Levofloxacin & 0.032 & 3.22 & 24.72 & 33.41 & 28.72 & 17.29 & 39.46 & 34.74 & 37.03 & 119.4 \\
{[}Medications{]} Thiamine & 0.023 & 2.33 & 20.10 & 27.07 & 28.47 & 15.60 & 34.73 & 31.13 & 34.85 & 118.8 \\
{[}Blood Products/Colloids{]} Platelets & 0.028 & 2.83 & 25.27 & 28.77 & 28.92 & 13.46 & 33.92 & 29.73 & 35.13 & 117.5 \\
{[}Medications{]} Acetaminophen-IV & 0.028 & 2.75 & 35.08 & 44.09 & 34.62 & 17.31 & 49.88 & 44.36 & 48.76 & 116.3 \\
{[}Medications{]} Epinephrine & 0.010 & 0.98 & 37.63 & 35.58 & 34.90 & 30.36 & 46.00 & 46.20 & 48.54 & 113.6 \\
{[}Blood Products/Colloids{]} OR Packed RBC Intake & 0.007 & 0.73 & 11.31 & 6.44 & 8.18 & 9.48 & 3.10 & 3.85 & 5.72 & 111.7 \\
{[}Medications{]} Lorazepam (Ativan) & 0.079 & 7.92 & 26.60 & 40.82 & 32.62 & 22.42 & 40.67 & 38.71 & 40.49 & 111.3 \\
{[}Medications{]} Nitroglycerin & 0.026 & 2.57 & 27.10 & 35.43 & 35.27 & 24.33 & 42.35 & 38.73 & 43.95 & 110.2 \\
{[}Medications{]} Folic Acid & 0.013 & 1.27 & 21.47 & 22.99 & 30.14 & 13.08 & 31.49 & 29.44 & 27.48 & 90.2 \\
{[}Blood Products/Colloids{]} Packed Red Blood Cells & 0.113 & 11.28 & 31.10 & 34.10 & 34.62 & 24.19 & 31.88 & 29.58 & 31.82 & 80.3 \\
{[}Medications{]} Magnesium Sulfate & 0.209 & 20.90 & 39.20 & 40.15 & 37.32 & 31.33 & 39.94 & 39.56 & 40.95 & 79.5 \\
{[}Medications{]} Hydralazine & 0.082 & 8.19 & 35.02 & 46.67 & 41.25 & 29.86 & 45.20 & 45.81 & 47.66 & 75.1 \\
{[}Antibiotics{]} Ceftriaxone & 0.045 & 4.51 & 39.29 & 49.14 & 46.75 & 32.14 & 50.61 & 48.14 & 50.37 & 70.1 \\
{[}Medications{]} Metoprolol & 0.139 & 13.87 & 40.43 & 50.15 & 47.09 & 35.10 & 50.03 & 48.23 & 51.04 & 65.8 \\
{[}Medications{]} Nicardipine & 0.022 & 2.18 & 41.27 & 57.31 & 51.79 & 37.49 & 63.34 & 61.12 & 61.67 & 65.8 \\
{[}Blood Products/Colloids{]} OR Colloid Intake & 0.003 & 0.34 & 2.15 & 2.11 & 5.62 & 1.38 & 1.06 & 0.98 & 1.58 & 64.8 \\
{[}Medications{]} Morphine Sulfate & 0.105 & 10.46 & 35.70 & 43.69 & 45.22 & 33.67 & 46.60 & 45.87 & 48.92 & 63.5 \\
{[}Medications{]} Calcium Gluconate & 0.158 & 15.81 & 44.69 & 50.20 & 46.03 & 35.82 & 47.16 & 45.23 & 48.74 & 63.3 \\
{[}Medications{]} Hydromorphone (Dilaudid) & 0.092 & 9.23 & 46.13 & 51.50 & 51.07 & 40.62 & 54.24 & 56.20 & 57.49 & 61.0 \\
{[}Medications{]} Phenylephrine & 0.117 & 11.72 & 47.26 & 64.02 & 56.17 & 42.32 & 65.78 & 63.43 & 65.46 & 57.6 \\
{[}Medications{]} Vasopressin & 0.043 & 4.32 & 48.72 & 59.36 & 56.86 & 47.00 & 67.15 & 69.36 & 71.10 & 56.8 \\
{[}Antibiotics{]} Cefazolin & 0.042 & 4.24 & 50.23 & 59.91 & 57.90 & 53.46 & 61.63 & 65.61 & 67.93 & 52.5 \\
{[}Medications{]} KCL & 0.189 & 18.86 & 50.10 & 55.70 & 52.08 & 43.17 & 53.05 & 52.78 & 54.61 & 51.1 \\
{[}Blood Products/Colloids{]} OR Platelet Intake & 0.004 & 0.38 & 7.48 & 7.22 & 7.45 & 12.98 & 1.54 & 2.88 & 3.70 & 48.9 \\
{[}Medications{]} Dilantin & 0.031 & 3.06 & 47.67 & 56.09 & 57.09 & 46.39 & 60.16 & 57.70 & 61.10 & 48.6 \\
{[}Antibiotics{]} Acyclovir & 0.030 & 3.03 & 55.38 & 67.97 & 63.29 & 48.40 & 76.48 & 73.71 & 76.91 & 48.5 \\
{[}Medications{]} Furosemide & 0.264 & 26.37 & 53.38 & 60.51 & 58.29 & 48.56 & 60.23 & 60.32 & 62.30 & 47.0 \\
{[}Antibiotics{]} Ciprofloxacin & 0.100 & 9.95 & 51.48 & 62.13 & 62.14 & 49.77 & 62.75 & 60.89 & 62.85 & 44.1 \\
{[}Medications{]} Norepinephrine & 0.122 & 12.15 & 60.11 & 67.82 & 68.36 & 61.07 & 71.39 & 72.36 & 71.81 & 34.5 \\
{[}Medications{]} Propofol & 0.192 & 19.18 & 57.77 & 65.20 & 66.72 & 56.32 & 69.12 & 67.25 & 69.93 & 34.5 \\
{[}Medications{]} ACD-A Citrate & 0.038 & 3.80 & 66.52 & 65.68 & 70.10 & 65.86 & 74.22 & 76.98 & 80.93 & 33.9 \\
{[}Medications{]} Pantoprazole (Protonix) & 0.218 & 21.75 & 61.43 & 66.93 & 65.96 & 54.67 & 63.87 & 62.41 & 64.62 & 31.7 \\
{[}Antibiotics{]} Cefepime & 0.132 & 13.22 & 63.46 & 67.47 & 70.90 & 63.49 & 70.52 & 73.68 & 73.96 & 31.6 \\
{[}Medications{]} Potassium Chloride & 0.342 & 34.25 & 63.73 & 67.93 & 66.76 & 56.98 & 66.33 & 64.93 & 67.29 & 31.3 \\
{[}Medications{]} Fentanyl & 0.275 & 27.46 & 66.16 & 71.43 & 70.32 & 62.90 & 72.06 & 72.39 & 74.75 & 29.8 \\
{[}Medications{]} Famotidine (Pepcid) & 0.201 & 20.14 & 66.36 & 71.36 & 70.52 & 62.19 & 69.16 & 69.20 & 70.46 & 27.6 \\
{[}Antibiotics{]} Metronidazole & 0.111 & 11.08 & 65.55 & 71.70 & 74.28 & 65.56 & 74.88 & 73.43 & 75.71 & 26.4 \\
{[}Antibiotics{]} Vancomycin & 0.331 & 33.07 & 67.78 & 70.14 & 73.83 & 64.72 & 71.41 & 72.68 & 73.64 & 25.2 \\
{[}Blood Products/Colloids{]} OR Cell Saver Intake & 0.002 & 0.23 & 7.65 & 7.03 & 11.63 & 7.20 & 1.35 & 2.96 & 5.82 & 24.3 \\
{[}Medications{]} Heparin Sodium & 0.120 & 12.02 & 72.86 & 75.11 & 75.31 & 68.58 & 75.92 & 76.01 & 77.45 & 23.8 \\
{[}Medications{]} Midazolam (Versed) & 0.201 & 20.06 & 74.08 & 75.53 & 77.06 & 69.11 & 76.02 & 76.38 & 77.23 & 20.9 \\
{[}Antibiotics{]} Meropenem & 0.097 & 9.65 & 72.10 & 78.54 & 79.03 & 72.36 & 78.95 & 79.93 & 81.23 & 20.7 \\
\bottomrule
\end{longtable}
}

\scriptsize{
\begin{longtable}{@{}p{0.27\linewidth} | p{0.06\linewidth} | p{0.06\linewidth} | p{0.06\linewidth} | p{0.06\linewidth} | p{0.06\linewidth} | p{0.06\linewidth} | p{0.06\linewidth} | p{0.08\linewidth} | p{0.08\linewidth} | p{0.07\linewidth} @{}}
\caption{Performance of each procedure event, sorted by performance gain between Meta-Switch-Event vs. GRU-POP}\\
\toprule
\textbf{Procedure Event Types}   & \textbf{Freq.}    & \textbf{CNN} & \textbf{RE TAIN}  & \textbf{GRU-POP} & \textbf{Sub-pop Adap} & \textbf{Self Adap} & \textbf{Com bined Adap}  & \textbf{Meta-Switch} & \textbf{Meta-Switch-Event} & \textbf{Gain (GRU-POP vs. Meta-Switch-Event.}\\ \midrule
Abdominal X-Ray & 0.008 & 0.85 & 1.17 & 1.13 & 1.45 & 1.03 & 3.28 & 2.97 & 2.40 & 896.9 \\
Interventional Radiology & 0.010 & 0.95 & 1.23 & 1.65 & 1.40 & 1.36 & 1.55 & 1.37 & 1.59 & 835.6 \\
Endoscopy & 0.005 & 0.54 & 1.52 & 2.33 & 2.12 & 1.39 & 2.86 & 4.15 & 3.44 & 761.4 \\
X-ray & 0.027 & 2.69 & 3.55 & 3.97 & 3.93 & 3.37 & 5.18 & 5.56 & 6.05 & 557.9 \\
Stool Culture & 0.020 & 1.99 & 3.09 & 3.55 & 3.66 & 2.83 & 4.42 & 4.12 & 4.63 & 538.7 \\
Trans Esophageal Echo & 0.007 & 0.74 & 2.42 & 1.91 & 2.42 & 1.57 & 1.60 & 1.38 & 1.78 & 435.4 \\
EEG & 0.012 & 1.15 & 3.59 & 3.87 & 3.87 & 2.82 & 3.93 & 4.23 & 4.73 & 421.5 \\
Family meeting held & 0.013 & 1.34 & 2.87 & 4.65 & 4.65 & 2.77 & 5.55 & 5.02 & 6.01 & 405.1 \\
Nasal Swab & 0.010 & 0.97 & 3.53 & 1.95 & 3.12 & 2.13 & 1.29 & 1.55 & 1.65 & 334.1 \\
Bronchoscopy & 0.027 & 2.74 & 4.95 & 5.73 & 5.85 & 4.17 & 5.39 & 5.06 & 6.07 & 307.1 \\
Transthoracic Echo & 0.030 & 2.98 & 6.24 & 5.93 & 6.18 & 5.53 & 5.69 & 4.86 & 5.58 & 290.5 \\
Sputum Culture & 0.039 & 3.89 & 5.92 & 7.04 & 7.75 & 5.84 & 7.68 & 7.47 & 9.15 & 284.5 \\
Sheath & 0.003 & 0.34 & 9.28 & 27.62 & 13.70 & 16.94 & 28.01 & 24.89 & 31.71 & 246.8 \\
Urine Culture & 0.058 & 5.84 & 6.82 & 8.60 & 8.40 & 6.76 & 7.64 & 7.30 & 9.71 & 235.1 \\
Pan Culture & 0.022 & 2.20 & 4.27 & 7.72 & 6.26 & 4.80 & 4.25 & 5.03 & 5.59 & 222.0 \\
Ultrasound & 0.044 & 4.38 & 6.79 & 8.43 & 8.29 & 5.44 & 6.39 & 5.83 & 7.23 & 220.7 \\
Intubation & 0.028 & 2.75 & 4.62 & 6.89 & 8.06 & 4.72 & 4.33 & 3.56 & 6.53 & 202.2 \\
Blood Cultured & 0.082 & 8.23 & 11.91 & 13.87 & 13.17 & 11.07 & 13.12 & 13.30 & 14.96 & 200.8 \\
Family updated by MD & 0.021 & 2.08 & 7.37 & 13.29 & 12.42 & 5.29 & 13.44 & 12.15 & 13.47 & 197.7 \\
Family updated by RN & 0.036 & 3.57 & 21.58 & 28.13 & 20.87 & 9.89 & 27.66 & 22.20 & 27.43 & 168.0 \\
EKG & 0.078 & 7.80 & 14.07 & 13.04 & 13.40 & 11.09 & 12.37 & 12.34 & 13.32 & 167.9 \\
OR Sent & 0.024 & 2.45 & 7.56 & 9.37 & 7.93 & 5.99 & 5.35 & 5.32 & 6.53 & 165.3 \\
Magnetic Resonance Imaging & 0.016 & 1.63 & 5.43 & 5.26 & 6.28 & 3.92 & 3.42 & 2.96 & 3.70 & 157.8 \\
OR Received & 0.025 & 2.54 & 8.10 & 10.01 & 9.42 & 9.06 & 5.87 & 6.32 & 7.60 & 143.1 \\
CT scan & 0.078 & 7.78 & 17.55 & 17.97 & 18.49 & 12.36 & 14.13 & 13.51 & 15.04 & 124.1 \\
Extubation & 0.071 & 7.14 & 15.35 & 15.46 & 19.62 & 16.44 & 7.76 & 13.41 & 13.24 & 121.5 \\
Hemodialysis & 0.027 & 2.73 & 29.79 & 35.38 & 33.30 & 26.06 & 42.50 & 42.18 & 45.35 & 114.5 \\
Non-invasive Ventilation & 0.025 & 2.54 & 22.63 & 44.24 & 39.91 & 34.32 & 54.24 & 55.21 & 57.67 & 100.7 \\
Chest X-Ray & 0.277 & 27.73 & 43.93 & 46.46 & 45.38 & 37.72 & 41.91 & 41.66 & 43.40 & 64.9 \\
PA Catheter & 0.026 & 2.63 & 47.85 & 71.10 & 63.99 & 53.52 & 70.65 & 70.51 & 71.85 & 39.4 \\
Chest Tube Removed & 0.011 & 1.06 & 14.81 & 18.56 & 21.22 & 13.12 & 6.27 & 10.19 & 11.19 & 36.7 \\
16 Gauge & 0.056 & 5.57 & 60.34 & 64.81 & 68.39 & 60.64 & 64.30 & 67.66 & 68.97 & 34.3 \\
22 Gauge & 0.105 & 10.47 & 60.96 & 64.80 & 70.25 & 63.67 & 68.50 & 70.35 & 70.59 & 31.2 \\
CCO PAC & 0.020 & 2.01 & 67.69 & 76.08 & 77.98 & 72.24 & 74.37 & 79.00 & 80.35 & 21.4 \\
Dialysis - CRRT & 0.051 & 5.10 & 78.61 & 76.03 & 80.54 & 73.32 & 82.66 & 83.91 & 86.04 & 19.1 \\
18 Gauge & 0.229 & 22.88 & 77.19 & 76.84 & 81.25 & 75.86 & 73.38 & 78.80 & 77.65 & 18.7 \\
Cordis/Introducer & 0.062 & 6.16 & 78.56 & 83.52 & 84.65 & 79.22 & 84.46 & 87.90 & 87.99 & 14.5 \\
Indwelling Port (PortaCath) & 0.024 & 2.37 & 90.13 & 98.35 & 88.72 & 81.03 & 97.07 & 96.40 & 97.33 & 11.5 \\
20 Gauge & 0.391 & 39.08 & 83.66 & 84.61 & 88.05 & 84.64 & 82.91 & 87.13 & 87.17 & 10.5 \\
Dialysis Catheter & 0.104 & 10.41 & 90.05 & 91.86 & 92.39 & 89.24 & 92.76 & 92.81 & 92.63 & 6.8 \\
PICC Line & 0.273 & 27.32 & 92.86 & 91.98 & 93.51 & 92.10 & 92.88 & 93.32 & 93.34 & 5.9 \\
Arterial Line & 0.519 & 51.92 & 93.17 & 93.71 & 94.71 & 92.60 & 92.68 & 94.34 & 94.92 & 4.8 \\
Invasive Ventilation & 0.501 & 50.08 & 94.95 & 95.81 & 95.95 & 94.97 & 95.56 & 95.90 & 96.31 & 3.6 \\
\bottomrule
\end{longtable}
}

\scriptsize{
\begin{longtable}{@{}p{0.27\linewidth} | p{0.06\linewidth} | p{0.06\linewidth} | p{0.06\linewidth} | p{0.06\linewidth} | p{0.06\linewidth} | p{0.06\linewidth} | p{0.06\linewidth} | p{0.08\linewidth} | p{0.08\linewidth} | p{0.07\linewidth} @{}}
\caption{Performance of each physiological event, sorted by performance gain between Meta-Switch-Event vs. GRU-POP}\\
\toprule
\textbf{Physiological Event Types}   & \textbf{Freq.}    & \textbf{CNN} & \textbf{RE TAIN}  & \textbf{GRU-POP} & \textbf{Sub-pop Adap} & \textbf{Self Adap} & \textbf{Com bined Adap}  & \textbf{Meta-Switch} & \textbf{Meta-Switch-Event} & \textbf{Gain (GRU-POP vs. Meta-Switch-Event.}\\ \midrule
Cardiovascular: LUE Color & 0.544 & 54.38 & 76.35 & 78.24 & 77.78 & 71.63 & 76.94 & 77.10 & 80.25 & 19.6 \\
Cardiovascular: RUE Color & 0.547 & 54.74 & 76.65 & 78.38 & 77.91 & 72.11 & 77.01 & 77.31 & 80.43 & 19.5 \\
Cardiovascular: LLE Color & 0.560 & 55.96 & 77.05 & 78.90 & 78.38 & 71.99 & 77.33 & 77.69 & 80.72 & 19.0 \\
Cardiovascular: LUE Temp & 0.555 & 55.51 & 77.03 & 79.30 & 78.80 & 73.04 & 78.01 & 78.16 & 81.11 & 18.6 \\
Cardiovascular: RLE Color & 0.563 & 56.30 & 77.24 & 79.00 & 78.79 & 72.44 & 77.36 & 78.08 & 80.92 & 18.4 \\
Cardiovascular: RUE Temp & 0.558 & 55.82 & 77.32 & 79.48 & 78.97 & 73.44 & 78.10 & 78.30 & 81.25 & 18.4 \\
Cardiovascular: LLE Temp & 0.571 & 57.11 & 77.62 & 79.79 & 79.34 & 73.36 & 78.29 & 78.61 & 81.35 & 18.0 \\
Cardiovascular: RLE Temp & 0.574 & 57.43 & 77.89 & 79.93 & 79.69 & 73.83 & 78.30 & 78.89 & 81.44 & 17.5 \\
Routine Vital Signs: Arterial Blood Pressure diastolic & 0.496 & 49.60 & 91.64 & 91.64 & 93.93 & 91.59 & 90.65 & 92.75 & 93.36 & 5.6 \\
Routine Vital Signs: Arterial Blood Pressure systolic & 0.496 & 49.60 & 91.66 & 91.64 & 93.94 & 91.59 & 90.65 & 92.75 & 93.36 & 5.6 \\
Routine Vital Signs: Arterial Blood Pressure mean & 0.497 & 49.75 & 91.87 & 91.79 & 94.02 & 91.53 & 90.70 & 92.79 & 93.42 & 5.5 \\
Respiratory: Peak Insp. Pressure & 0.496 & 49.64 & 93.92 & 94.22 & 95.13 & 93.46 & 92.67 & 93.92 & 94.71 & 4.1 \\
Respiratory: PEEP set & 0.504 & 50.42 & 94.57 & 94.92 & 95.67 & 94.05 & 93.39 & 94.55 & 95.24 & 3.7 \\
Respiratory: Inspired O2 Fraction & 0.642 & 64.20 & 95.31 & 95.85 & 96.11 & 95.07 & 94.80 & 96.13 & 96.32 & 3.3 \\
Routine Vital Signs: Temperature Fahrenheit & 0.941 & 94.15 & 98.05 & 97.90 & 98.56 & 97.81 & 98.45 & 98.56 & 98.62 & 1.3 \\
Respiratory: Respiratory Rate & 0.994 & 99.39 & 99.70 & 99.92 & 99.86 & 99.86 & 99.92 & 99.91 & 99.92 & 0.1 \\
Respiratory: O2 saturation pulseoxymetry & 0.992 & 99.23 & 99.81 & 99.81 & 99.93 & 99.77 & 99.79 & 99.78 & 99.85 & 0.1 \\
Routine Vital Signs: Heart Rhythm & 0.995 & 99.47 & 99.88 & 99.92 & 99.97 & 99.90 & 99.94 & 99.90 & 99.91 & 0.0 \\
Routine Vital Signs: Heart Rate & 0.995 & 99.55 & 99.91 & 99.96 & 99.98 & 99.97 & 99.96 & 99.96 & 99.97 & 0.0 \\
\bottomrule
\end{longtable}
}
\end{document}